\pdfoutput=1
\documentclass[11pt]{article}

\usepackage[final]{acl}

\usepackage{times}
\usepackage{latexsym}
\usepackage{booktabs}
\usepackage{multirow} 
\usepackage{booktabs}
\usepackage{graphicx}
\usepackage{algorithm}
\usepackage{algpseudocode}
\usepackage{amsmath, amssymb, stmaryrd}
\usepackage{tipa}
\usepackage{graphicx}
\usepackage{fontawesome5}
\usepackage[most]{tcolorbox}

\newtcolorbox{promptbox}{
fontupper=\footnotesize,
colback=green!12,
colframe=black,
boxrule=0.5pt,
arc=5pt,
left=4pt,
right=4pt,
top=3pt,
bottom=3pt,
enhanced jigsaw,
}

\usepackage[T1]{fontenc}

\usepackage[utf8]{inputenc}
\usepackage{tipa}
\DeclareUnicodeCharacter{026A}{\textipa{I}}
\DeclareUnicodeCharacter{03B8}{\textipa{T}}
\DeclareUnicodeCharacter{0279}{\textipa{R}}       
\DeclareUnicodeCharacter{028C}{\textipa{2}}       
\DeclareUnicodeCharacter{0361}{}                  
\DeclareUnicodeCharacter{02DE}{\textsuperscript{r}} 
\DeclareUnicodeCharacter{026A}{\textipa{I}}      
\DeclareUnicodeCharacter{025B}{\textipa{E}}      
\DeclareUnicodeCharacter{00E6}{\textipa{\ae}}    
\DeclareUnicodeCharacter{0251}{\textipa{A}}      
\DeclareUnicodeCharacter{0254}{\textipa{O}}      
\DeclareUnicodeCharacter{028A}{\textipa{U}}      
\DeclareUnicodeCharacter{025D}{\textrhookrevepsilon} 
\DeclareUnicodeCharacter{0259}{\textipa{@}}      

\DeclareUnicodeCharacter{0283}{\textipa{S}}      
\DeclareUnicodeCharacter{0292}{\textipa{Z}}      
\DeclareUnicodeCharacter{014B}{\textipa{N}}      
\DeclareUnicodeCharacter{00F0}{\textipa{D}}      
\DeclareUnicodeCharacter{03B8}{\textipa{T}}      

\DeclareUnicodeCharacter{02A7}{\textipa{\t{tS}}} 
\DeclareUnicodeCharacter{02A4}{\textipa{\t{dZ}}} 

\DeclareUnicodeCharacter{0261}{\textipa{g}}      
\usepackage{amsmath}

\usepackage{microtype}

\usepackage{inconsolata}

\usepackage{graphicx}

\title{PEDBench: Can Speech Models Detect Phonetic Errors?}

\author{
 \textbf{Zhonghao Shi\textsuperscript{*,1}},
 \textbf{Zackary Rackauckas\textsuperscript{*,1}},
 \textbf{Jasmine Yip\textsuperscript{2}},
 \textbf{Savannah Halpern\textsuperscript{3}},
\\
 \textbf{Simon Campos\textsuperscript{3}},
 \textbf{Lauren Brace\textsuperscript{3}},
 \textbf{Emma Yang\textsuperscript{4}},
 \textbf{Tiantian Feng\textsuperscript{2}},
\\
 \textbf{Xuan Shi\textsuperscript{2}},
 \textbf{Shrikanth Narayanan\textsuperscript{2}},
 \textbf{Teresa Satterfield\textsuperscript{3}},
 \textbf{Ying Xu\textsuperscript{1}}
\\[0.8em]
\parbox{0.9\textwidth}{%
    \centering
    \textsuperscript{1}Harvard University,
    \textsuperscript{2}University of Southern California,
    \textsuperscript{3}University of Michigan,
    \textsuperscript{4}Columbia University \\
    {\small
    \textbf{Correspondence:}
    \{zhonghao\_shi, zackary\_rackauckas, yingxu\}@gse.harvard.edu
    }
}
}

\begin{document}
\maketitle

\begingroup
\renewcommand{\thefootnote}{*}
\footnotetext[1]{These authors contributed equally.}
\endgroup

\begin{abstract}

Phonetic error detection (PED) aims to identify phoneme-level substitutions, deletions, and insertions and is essential for educational and clinical speech assessment and intervention. Despite recent advances in speech models for phoneme recognition, their ability to reliably detect fine-grained phonetic errors remains unclear because of a lack of high-quality evaluation data. To address this gap, we introduce \textbf{PEDBench}, the first PED benchmark spanning a continuum from natural to controlled speech. It comprises real-world recordings of spoken errors made by children (N=334), scripted error productions by human actors (N=1{,}188), and synthetic error productions generated using phone-conditioned text-to-speech (N=1{,}083). All phonetic annotations were transcribed and verified by two phonetically trained annotators. Our evaluation of 13 state-of-the-art speech models shows that they fail to reliably detect phonetic errors, even under controlled conditions with acted and synthetic speech, where acoustic variability is reduced. The models also exhibit low precision, frequently misclassifying correctly pronounced phonemes as errors. Our speech-language models, fine-tuned on synthetic data, outperform commercial and open-source baselines on word-level PED, although a sim-to-real domain gap remains. We plan to publicly release our code and recipes, along with our acted and synthetic datasets, to support the development of more accurate PED systems.

\end{abstract}

\section{Introduction}

Automatic phoneme-level pronunciation assessment aims to detect phonetic errors in speech, with applications ranging from literacy education to clinical speech-language pathology. The need for scalable assessment and support is substantial. Recent national assessment results show that 40\% of U.S. fourth-grade students perform below basic level in reading~\cite{nces2024naep}. In parallel, a national survey found that 1 in 13 U.S. children ages 3--17 experienced a voice, speech, or language disorder during the preceding 12 months, yet more than 40\% of these children received no treatment during that period~\cite{black2015communication}. Prior work attributes these gaps, in part, to limited access to accurate early assessment and affordable, effective intervention~\cite{morgan2016receives}. Speech models that reliably detect phonetic errors could enable tools that expand access to early screening and supplement educational and clinical interventions.

\begin{figure}[t]
\centering
\includegraphics[width=\columnwidth]{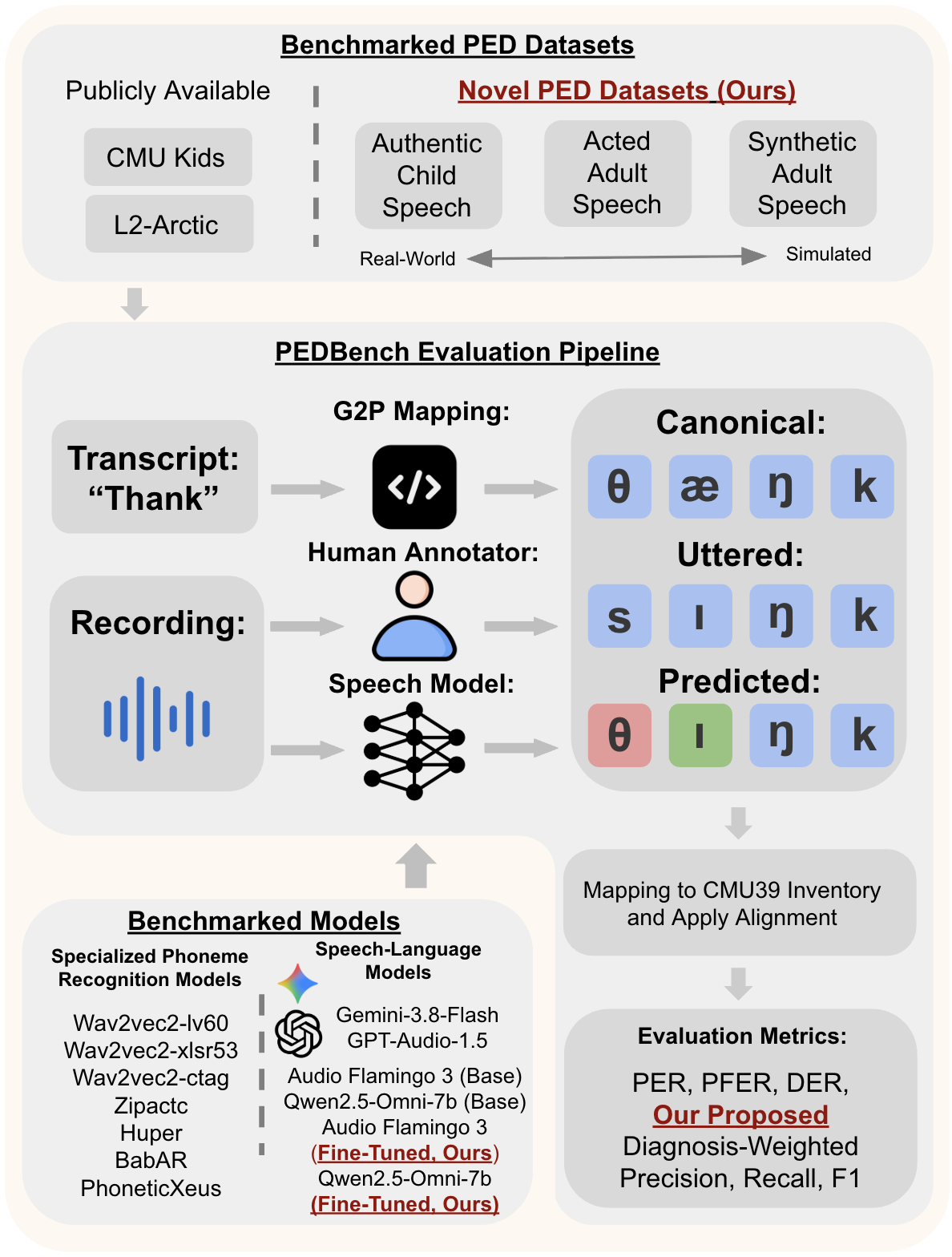}
\caption{Overview of the inference and evaluation pipeline for phonetic error detection.}
\label{fig:overview}
\end{figure}

A substantial body of work has investigated PED and related tasks, including mispronunciation detection and diagnosis (MDD)~\cite{lounis2024mispronunciation}, disfluency diagnosis~\cite{lian2024ssdm}, and automatic speech assessment~\cite{el2023automatic}. Nevertheless, whether state-of-the-art (SOTA) speech models can detect phonetic errors reliably enough for real-world deployment remains unclear. We identify three key limitations that hinder progress. First, \textit{inconsistent ground-truth annotations}: our analysis reveals inconsistencies in human phoneme transcriptions in existing PED datasets, undermining the reliability and validity of evaluations. Second, \textit{limited evaluation of SOTA models}: prior work has not yet evaluated SOTA general-purpose phoneme recognition models and speech-language models for PED, while many specialized PED models remain closed-source, limiting their downstream use. Third, \textit{inadequate evaluation metrics}: commonly reported metrics, including phoneme and phoneme-feature error rates for PR and precision, recall, and F1 scores for MDD, may not adequately characterize a model's ability to detect phonetic errors.

To address these limitations, we introduce \textbf{PEDBench}, the first benchmark for systematically evaluating state-of-the-art (SOTA) speech models on PED. Our contributions are threefold:

\begin{enumerate}
\item \textbf{Novel PED datasets with verified ground-truth transcriptions:} We curate three novel datasets: 1) natural child speech with errors; 2) scripted error productions by adult actors; and 3) synthetic adult speech with errors. Together, these datasets span a continuum from naturalistic, real-world speech to controlled, simulated speech, enabling evaluation across varying levels of acoustic variability and task difficulty. All phoneme-level transcriptions are independently produced or verified by two phonetically trained annotators.

\item \textbf{Evaluation of SOTA speech models:} We evaluate 8 specialized phoneme recognition (PR) models, 1 specialized commercial PED model, 2 commercial speech-language models (SLMs), and 2 open-source SLMs on public adult and child speech datasets and our novel PED datasets. We report both PED and phoneme recognition metrics, including our proposed diagnosis-weighted precision, recall, and F1 scores, which require both correct error detection and correct diagnosis rather than detection alone. Our results show that SOTA speech models still fail to detect even simple phonetic errors from word-level speech, under controlled conditions without noise or other sources of acoustic variability.

\item \textbf{Supervised fine-tuning of SLMs with synthetic speech for PED:} We show that LoRA fine-tuning of Qwen2.5-Omni-7B on TIMIT provides a strong baseline for PED. We further demonstrate that training with synthetic speech with errors generated from phoneme-controlled TTS further improves Qwen2.5-Omni-7B's performance, enabling it to outperform all evaluated baseline models. These results highlight the potential of synthetic speech to scale training data for robust PED.
\end{enumerate}


\section{Related Work}

\subsection{Phonetic Error Detection}

Prior work has investigated phonetic error detection (PED) in contexts such as mispronunciation detection and diagnosis (MDD)~\cite{lounis2024mispronunciation}, disfluency diagnosis~\cite{lian2024ssdm}, and automatic speech assessment~\cite{el2023automatic}. Most prior studies formulate PED as a phoneme recognition problem~\cite{leung2019cnn}: a phoneme recognizer is trained using connectionist temporal classification (CTC) loss, and the predicted phoneme sequence is aligned with the canonical sequence to identify errors. More recently, \citet{zhou2025towards} proposed a multi-task objective that incorporates phoneme similarity, while \citet{li2026k} introduced the Kids-Weighted Finite-State Transducer (K-WFST). However, these recent specialized PED models~\cite{zhou2025towards, li2026k} remain closed-source. More broadly, PED research remains fragmented and lacks a standardized benchmark for measuring progress.

\subsection{State-of-the-art Phoneme Speech Models}

Recent advances in speech processing have led to substantial improvements in phoneme recognition~\cite{bharadwaj2026prism}. However, these state-of-the-art specialized phoneme recognition (PR) models have not yet been systematically evaluated for PED. W2V2P-XLSR53 adapts self-supervised wav2vec~2.0 models through pretraining and fine-tuning on diverse data mixtures~\cite{xu2021simple}. ZIPA models~\cite{zhu2025zipa} use encoder–CTC architectures trained from scratch on multilingual data with consistency-regularized CTC (CR-CTC) loss. More recently, \citet{lavechin2026babar} introduced BabAR, a phoneme recognition model both pre-trained and fine-tuned on multi-lingual child speech.


Speech-language models (SLMs) are promising candidates for downstream PED applications such as reading instruction. Recent work~\cite{bharadwaj2026prism} has shown that both commercial and open-source SLMs achieve competitive PFERs on public phoneme recognition benchmarks, although they still lag behind specialized PR models. \citet{liu2026unlocking} explored training SLMs with a Whisper audio encoder and a language-model decoder for word-level mispronunciation detection. However, the PED capabilities of state-of-the-art SLMs remain underexplored.

\section{PEDBench}
\label{sec:PEDBench}
As shown in Figure~\ref{fig:overview}, PEDBench aims to establish the first standardized benchmark for evaluating state-of-the-art (SOTA) speech models on phonetic error detection (PED). The benchmark addresses a central question: \textit{Can SOTA speech models detect phonetic errors in controlled and real-world settings?} This section describes the details of the benchmark design.


\subsection{Benchmarked Datasets}

\subsubsection{Public Datasets}
We included two of the most commonly used public benchmarks in the PED and MDD literature: L2-ARCTIC-Perceived (N=3599)~\citep{zhao2018l2} and CMU-Kids (N=5180)~\citep{Mostow-1996-14274}. More details can be found in Appendix~\ref{sec:public-data}.

\subsubsection{Our Novel PED Datasets}
\label{subsubsec:novel-diagnostic-data}

Prior work has shown that phoneme production in real-world speech can be ambiguous. As illustrated in Figure~\ref{fig:spectrum}, speech sounds may fall near the phonetic boundary between two phoneme categories, leading to inconsistent transcriptions across human annotators and compromising evaluation validity. To reduce this annotation inconsistency, we adopt two approaches: (1) generating speech with phonetic errors using phoneme-controlled text-to-speech (TTS) models, and (2) recruiting native speakers to intentionally produce phonetic errors by following a script. For our real-world child speech data, we retain only samples for which two independent annotators provide identical phoneme transcriptions. Together, these efforts yield 3 novel PED datasets that enable the evaluation of SOTA speech models’ phonetic error detection performance in both controlled and real-world settings.


\paragraph{Synthetic Speech Dataset.}
\label{sec:synthetic-speech}
We generated synthetic speech data by embedding one phonetic substitution or deletion error into target words and synthesizing them in words, sentences, and passages using Amazon Polly text-to-speech (TTS), with pronunciation controlled through SSML \texttt{<phoneme>} tags. As shown in Table~\ref{tab:ipa-substitution-pairs}, we first identified two substitution errors for each phoneme and included the deletion error for each consonant. For consonants (N=22), we selected errors in consultation with a speech-language pathologist (SLP) collaborator. For vowels (N=14), we selected errors based on best available evidences we could find from prior speech-language pathology and literacy research~\cite{roepke2021vowel, cebrian2021easy, chung2014transcription}. We used 60 target words drawn from GFTA-3, a widely used assessment of speech-sound production~\citep{goldman1969goldman}. 

After generation, phonetically trained annotators reviewed the synthetic recordings in two independent rounds and flagged those that did not match their intended phoneme transcriptions. After excluding all flagged recordings, 1,141 recordings remained: 396 isolated words, 375 sentences, and 370 passages. From these recordings, we retained 361 complete matched sets, each containing an isolated-word recording and corresponding sentence and passage recordings with the same target word and phonetic error. The final synthetic evaluation dataset therefore comprised 1,083 utterances. This dataset enables us to evaluate models on objectively defined phonetic errors across different speech lengths while holding the target word and error constant. Separately, we also denote the 396 isolated synthetic words data as synthetic word (SW) training data used for supervised fine-tuning experiments.






\paragraph{Acted Speech Dataset.}
To bridge the gap between naturalistic and synthetic speech, we recruited three adult native English speakers with training in phonetics and asked them to intentionally produce scripted phonetic errors. The script matched the 396 SW training data in both target words and intended phonetic errors. After each speaker completed and confirmed their recordings, one of the other speakers reviewed and flagged any deviations from the script. Speakers re-recorded the flagged items until no recordings were flagged. The resulting acted-speech evaluation dataset comprised 1,188 utterances and also allows us to evaluate models on objectively defined phonetic errors, while serving as an intermediate evaluation condition between authentic child speech and fully synthetic speech.

\paragraph{Child Speech Dataset.}
\label{sec:child-speech}
We also collected child speech containing naturally occurring phonetic errors. We elicited word- and sentence-level speech from 199 typically developing Spanish--English bilingual children aged 3--7 years using a picture-naming task with 33 target words and three target sentences. Following data collection, two phonetically trained annotators independently transcribed 3,574 utterances at the phoneme level. We retained only recordings that both annotators identified as containing phonetic errors and for which they agreed exactly on the realized phoneme sequence. The resulting child-speech evaluation dataset comprised 334 utterances with high-confidence error labels from 147 children: 323 word recordings and 11 sentence recordings. In addition, we also identified all the error types in this child-speech evaluation dataset, and generated matching synthetic speech data (N=154). After combining it with the SC training dataset described in Section~\ref{sec:synthetic-speech}, we have the synthetic combined (SC) training dataset (N=550). More details can be found in Appendix~\ref{sec:ALMSpec-data-splits}

\subsection{Evaluation Pipeline}
\label{subsec:evaluation-pipeline}

Figure~\ref{fig:overview} illustrates the phonetic error detection (PED) pipeline used in PEDBench, which follows the standard approach in the mispronunciation detection and diagnosis (MDD) literature. For each speech recording, we obtain the canonical phoneme sequence $C$ by applying a grapheme-to-phoneme (G2P) conversion to its textual transcript. Human annotators provide the realized phoneme sequence $U$, while the speech model processes the recording to produce the predicted phoneme sequence $\hat{U}$. Our pipeline is implemented on top of the PRiSM toolkit \cite{bharadwaj2026prism}, reusing its inference recipes and phonetic-feature distance, and extending it with the CMU-39 projection (Appendix \ref{sec:cmu39}), MDD alignment (Appendix \ref{sec:mdd-alignment}), and PED metrics described below.

A challenge in prior work is that datasets and speech models use different phoneme inventories. Multilingual models, in particular, may emit IPA symbols outside the English phoneme inventory. PEDBench therefore standardizes all three sequences by projecting them onto the 39-phoneme inventory of the CMU Pronouncing Dictionary~\cite{cmudict}, represented in IPA, yielding $C^{39}$, $U^{39}$, and $\hat{U}^{39}$. Further details of phoneme standardization are provided in Appendix \ref{sec:cmu39}. Because these sequences generally differ in length, PEDBench first aligns them on a common grid following the hierarchical MDD evaluation framework~\citep{li2016mispronunciation}. Implementation details are provided in Appendix~\ref{sec:mdd-alignment}.

\subsection{Evaluation Metrics}
\label{subsec:metrics}

\paragraph{Phoneme-recognition metrics.}

PEDBench reports the standard phoneme error rate (PER) and phoneme feature error rates (PFER). Detailed definitions of these metrics are provided in the appendix~\ref{sec:appendix-metrics}.


\paragraph{PED metrics.}

\label{sec:alignment-metric}
\begin{figure}[t]
    \centering
    \includegraphics[width=\linewidth]{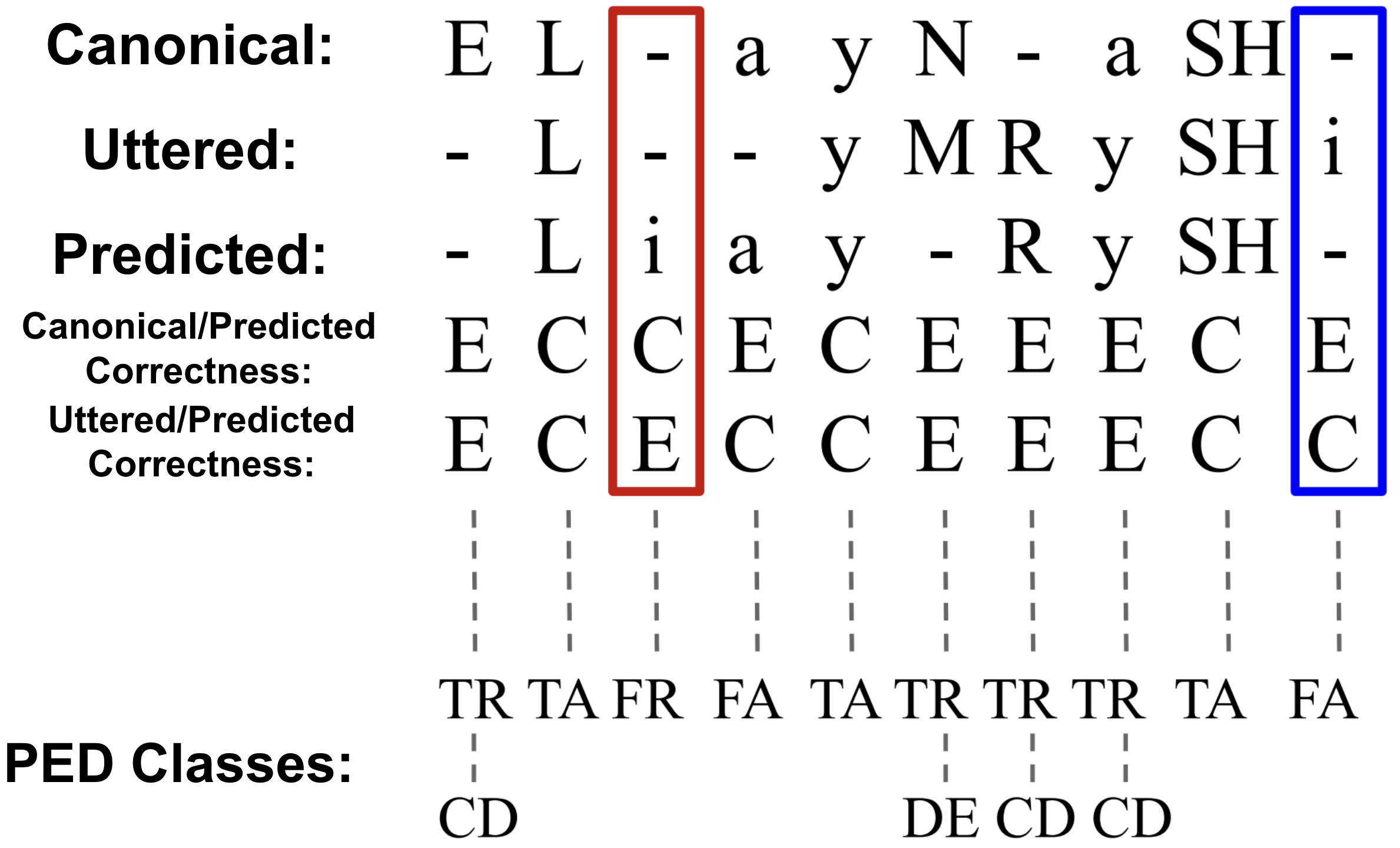}
    \caption{
    Phonetic Error Detection Alignment and Metric Calculation Rules~\cite{gelin2023comparing}.
    }
    \label{fig:alignment-metric}
\end{figure}


As illustrated in Figure~\ref{fig:alignment-metric}, each position in the alignment grid receives two correctness labels: canonical versus uttered (whether the phone was mispronounced) and canonical versus predicted (whether the model rejected the canonical phone). Together, these labels classify each position as a true acceptance (TA), false rejection (FR), false acceptance (FA), or true rejection (TR). Each TR is further classified as a correct diagnosis (CD) if the predicted phone matches the uttered phone, or a diagnostic error (DE) otherwise. From these counts, we compute the false rejection rate (FRR), false acceptance rate (FAR), overall diagnostic error rate (DER), and \emph{diagnosis-weighted} (DW) precision, recall, and F1. Standard MDD precision and recall use $\mathrm{TR}$ in the numerator, crediting a model for detecting an error even when its diagnosis is incorrect. Our DW variants instead use the stricter numerator $\mathrm{CD}$, awarding credit only when the model both detects the error and correctly identifies the realized phone, which is more aligned with real-world use needs.

\section{Benchmarked Models}

PEDBench evaluated 8 specialized phoneme recognition models, 1 specialized PED model, and 4 speech-language models (SLMs). 

\subsection{Specialized Phoneme Recognition Models}
\label{subsec:pr-models}

We evaluate 8 specialized phoneme-recognition (PR) models from 6 model families. From the wav2vec 2.0 family, we include its three top-performing systems: MultiIPA, W2V2P-LV60, and W2V2P-XLSR53~\citep{xu2021simple}. We also include ZIPA-CTC and ZIPA-CTC-NS, two Zipformer-based recognizers~\citep{zhu2025zipa}. We additionally include three recently released models: HuPER~\citep{guo2026huper}, PhoneticXeus~\citep{bharadwaj2026empirical}, and BabAR~\citep{lavechin2026babar}. 

\subsection{Specialized PED Model}
\label{subsec:ped-models}

We also evaluated the Azure's Pronunciation Assessment model~\cite{microsoft_pronunciation_assessment}, a specialized model designed to assess speech pronunciation on both the word and phoneme level that can be configured to perform phone recognition.


\subsection{Speech-Language Models}

Speech-language models (SLMs) are well suited to PED because their in-context learning capabilities support adaptation to applications with varying criteria for correct pronunciation. Their ability to generate text also supports downstream tasks such as reading instruction.

In PEDBench, we evaluate 2 proprietary SLMs, Gemini-3.8-Flash and GPT-Audio-1.5, and 2 open-weight SLMs, Qwen2.5-Omni-7B~\citep{Qwen2.5-Omni} and Audio Flamingo 3 (AF3)~\citep{ghosh2026audio}. We selected these models for their strong performance on recent audio-language benchmarks. To adapt Qwen2.5-Omni-7B and AF3 to phoneme recognition, we performed supervised fine-tuning using the training split of the TIMIT dataset~\cite{garofolo1993darpa}. Further training details are provided in the appendix~\ref{sec:ALMSpec}. We prompt each model to transcribe the input audio as a phoneme sequence using the 39-phoneme inventory of the CMU Pronouncing Dictionary~\cite{cmudict}. The system and user prompts are shown in Figure~\ref{fig:ALMSpec-prompt}.


\section{Fine-Tuning Experiments with Synthetic Data}

A core challenge in PED is collecting large-scale real-world training data covering diverse phonetic errors, particularly rare ones~\cite{macklin2020re}. Synthetic speech offers a promising way to scale training data while covering both common and rare error types. In this work, we ask the following question: \textit{Can synthetic speech data be used to fine-tune speech models for better PED?}

To answer this question, we fine-tuned three models---W2V2P-XLSR53, Qwen2.5-Omni-7B, and Audio Flamingo 3 (AF3)---and evaluated them on our new child-speech and acted-speech datasets. In downstream tasks such as reading assessment, common speech errors for each target word are often known. This information can guide the generation of synthetic training data to improve detection of these errors. As described in Sections~\ref{sec:synthetic-speech} and~\ref{sec:child-speech}, we constructed synthetic word (SW) training data covering the error types in the acted-speech evaluation dataset, and synthetic combined (SC) training data additionally covering those in the child-speech evaluation dataset. We fine-tuned each model separately on SW and SC, adding TIMIT to the training mixtures for Qwen2.5-Omni-7B and AF3. Further implementation details are provided in Section~\ref{sec:ALMSpec}.

\section{Results}

\subsection{Can SOTA speech models detect phonetic errors from L2-Arctic and CMU-Kids?}

\begin{table*}[!t]
  \centering
  \small
  \setlength{\tabcolsep}{4pt}
  \renewcommand{\arraystretch}{1.08}
  \resizebox{\textwidth}{!}{%
  \begin{tabular}{llrrrrrrrr}
  \toprule
  \textbf{Dataset} & \textbf{Model}
  & \multicolumn{2}{c}{\textbf{PR Metrics}}
  & \multicolumn{6}{c}{\textbf{PED Metrics}} \\
  \cmidrule(lr){3-4} \cmidrule(lr){5-10}
  & & \textbf{PFER $\downarrow$} & \textbf{PER $\downarrow$}
  & \textbf{FRR $\downarrow$} & \textbf{FAR $\downarrow$}
  & \textbf{DER $\downarrow$} & \textbf{Prec. $\uparrow$}
  & \textbf{Rec. $\uparrow$} & \textbf{F1 $\uparrow$} \\
  \midrule
  \multirow{16}{*}{L2-ARCTIC}
  & W2V2P-LV60       & 10.66 & 23.76 & 10.73 & 56.55 & 21.66 & 30.32 & 28.13 & 29.19 \\
  & W2V2P-XLSR53     & 11.42 & 26.35 & 15.33 & 44.01 & 24.02 & 28.52 & 36.06 & 31.85 \\
  & MultiIPA         & 19.21 & 58.00 & 48.57 & 16.19 & 49.97 & 13.37 & 43.17 & 20.42 \\
  & ZIPA-CTC         & 10.07 & 22.45 & 6.84 & 72.31 & 20.61 & 28.27 & 16.64 & 20.95 \\
  & ZIPA-CTC-NS      & 9.98 & 22.00 & 6.06 & 74.43 & 20.19 & 28.97 & 15.41 & 20.12 \\
  & HuPER            & 12.51 & 29.34 & 19.56 & 37.38 & 26.81 & 26.25 & 39.97 & 31.69 \\
  & PhoneticXeus     & 10.00 & 24.26 & 8.64 & 73.88 & 22.30 & 23.76 & 15.54 & 18.79 \\
  & BabAR            & 63.25 & 72.94 & 70.92 & \textbf{7.43} & 67.79 & 11.25 & \textbf{46.23} & 18.10 \\
  \addlinespace[2pt]
  & Azure Speech     & 10.67 & 24.03 & 11.04 & 55.43 & 22.01 & 29.33 & 27.87 & 28.58 \\
  \addlinespace[2pt]
  & Gemini-3.8-Flash & \textbf{9.58} & \textbf{20.54} & \textbf{5.30} & 72.45 & \textbf{19.00} & \textbf{37.17} & 19.13 & 25.26 \\
  & GPT-Audio-1.5    & 10.02 & 22.31 & 8.03 & 64.74 & 20.53 & 31.60 & 22.67 & 26.40 \\
  & Qwen2.5-Omni-7B  & 14.45 & 36.38 & 24.42 & 46.95 & 32.15 & 18.26 & 30.96 & 22.97 \\
  & Qwen$_{\mathrm{TIMIT}}$ \textbf{(ours)}
                     & 11.60 & 28.46 & 19.43 & 35.66 & 26.12 & 28.39 & 43.40 & \textbf{34.32} \\
  & Audio Flamingo 3 & 52.68 & 63.69 & 54.86 & 34.60 & 57.49 & 9.44 & 30.25 & 14.39 \\
  & AF3$_{\mathrm{TIMIT}}$ \textbf{(ours)}
                     & 12.24 & 28.95 & 18.59 & 43.10 & 26.63 & 26.05 & 36.82 & 30.51 \\
  \cmidrule(lr){2-10}
  & \textit{Human (100 random utts.)}
                     & 6.31 & 18.26 & 8.38 & 49.24 & 17.47 & 44.63 & 40.10 & 42.25 \\
  \midrule
  \multirow{13}{*}{CMU Kids}
  & W2V2P-LV60       & 8.39 & 17.55 & 12.70 & 23.30 & 16.46 & 45.04 & 64.11 & 52.91 \\
  & W2V2P-XLSR53     & 10.53 & 22.23 & 18.16 & 20.87 & 20.88 & 37.65 & 65.06 & 47.70 \\
  & MultiIPA         & 21.46 & 68.24 & 61.72 & \textbf{10.44} & 60.23 & 11.11 & 48.04 & 18.05 \\
  & ZIPA-CTC         & 8.39 & 18.04 & 13.40 & 22.22 & 16.90 & 44.19 & 65.02 & 52.62 \\
  & ZIPA-CTC-NS      & \textbf{7.86} & \textbf{16.72} & \textbf{11.90} & 23.67 & \textbf{15.66} & \textbf{47.05} & 64.88 & \textbf{54.55} \\
  & HuPER            & 12.95 & 26.90 & 23.95 & 16.20 & 25.16 & 32.96 & \textbf{68.54} & 44.52 \\
  & PhoneticXeus     & 10.01 & 24.30 & 19.75 & 22.00 & 22.83 & 34.07 & 61.27 & 43.79 \\
  & BabAR            & 19.85 & 37.66 & 34.91 & 15.94 & 34.85 & 24.49 & 65.47 & 35.65 \\
  & Qwen2.5-Omni-7B  & 12.94 & 30.43 & 24.70 & 25.52 & 27.69 & 27.22 & 56.29 & 36.70 \\
  & Qwen$_{\mathrm{TIMIT}}$ \textbf{(ours)}
                     & 12.53 & 24.65 & 21.14 & 19.02 & 23.25 & 34.80 & 65.94 & 45.56 \\
  & Audio Flamingo 3 & 49.44 & 56.53 & 51.97 & 38.86 & 52.94 & 12.73 & 42.02 & 19.54 \\
  & AF3$_{\mathrm{TIMIT}}$ \textbf{(ours)}
                     & 12.97 & 23.38 & 18.32 & 27.61 & 21.94 & 35.55 & 59.38 & 44.48 \\
  \cmidrule(lr){2-10}
  & \textit{Human (100 random utts.)}
                     & 4.68 & 9.55 & 5.25 & 20.92 & 9.20 & 63.72 & 68.92 & 66.22 \\
  \bottomrule
  \end{tabular}%
  }
  \caption{Benchmarked results on L2-ARCTIC and CMU Kids.}
  \label{tab:public-results}
  \end{table*}

As shown in Table~\ref{tab:public-results}, we evaluated all benchmarked models on L2-ARCTIC and CMU-Kids. Commercial models were excluded from CMU-Kids evaluation due to licensing requirements. This evaluation shows that even the best-performing models still fail to reliably detect phonetic errors in real-world, sentence-level speech, achieving only 34.32 F1 with TIMIT-fine-tuned Qwen2.5-Omni-7B on L2-ARCTIC and 54.55 F1 with ZIPA-CTC-NS on CMU-Kids. In addition, PFER may not adequately reflect PED performance. Although ZIPA-CTC-NS achieved an impressive PFER of 7.86 on CMU-Kids, its F1 of 54.55 remains unsatisfactory.

We also investigated whether dataset limitations may explain the poor performance. Qualitative inspection of both datasets suggested that some speech sounds in accented speech in L2-ARCTIC and child speech in CMU-Kids can be ambiguous and difficult to transcribe consistently, particularly vowels and voiceless consonants. To examine annotation consistency, we asked our phonetically trained annotators to reannotate a random sample of 100 utterances from each dataset. Evaluated against the original annotations, these human reannotations yielded F1 scores of 42.25 on L2-ARCTIC and 66.22 on CMU-Kids. This disagreement suggests that annotation uncertainty may also contribute to the low measured performance. These findings motivate curating PED datasets with more clearly defined and consistently verified phoneme transcriptions.

   \begin{table*}[!t]
  \centering
  \small
  \setlength{\tabcolsep}{4.5pt}
  \renewcommand{\arraystretch}{1.08}
  \resizebox{\textwidth}{!}{%
  \begin{tabular}{llrrrrrrrr}
  \toprule
  \textbf{Dataset} & \textbf{Model}
  & \multicolumn{2}{c}{\textbf{PR Metrics}}
  & \multicolumn{6}{c}{\textbf{PED Metrics}} \\
  \cmidrule(lr){3-4} \cmidrule(lr){5-10}
  & & \textbf{PFER $\downarrow$} & \textbf{PER $\downarrow$}
  & \textbf{FRR $\downarrow$} & \textbf{FAR $\downarrow$}
  & \textbf{DER $\downarrow$} & \textbf{Prec. $\uparrow$}
  & \textbf{Rec. $\uparrow$} & \textbf{F1 $\uparrow$} \\
  \midrule
  \multirow{18}{*}{Child}
  & W2V2P-LV60 & 27.53 & 62.38 & 50.53 & 7.38 & 49.71 & 24.92 & 52.20 & 33.73 \\
  & W2V2P-XLSR53 & 22.47 & 50.53 & 45.27 & 6.85 & 42.28 & 33.64 & 64.15 & 44.14 \\
  & W2V2P-XLSR53$_{\mathrm{SC}}$ \textbf{(ours)} & 21.30 & 47.27 & 42.42 & 7.38 & 39.10 & 36.74 & 68.19 & 47.75 \\
  & MultiIPA & 26.30 & 76.70 & 65.23 & \textbf{4.57} & 62.51 & 18.05 & 43.59 & 25.53 \\
  & ZIPA-CTC & 25.55 & 53.46 & 46.21 & 11.42 & 45.37 & 30.31 & 56.41 & 39.43 \\
  & ZIPA-CTC-NS & 24.47 & 53.46 & 46.06 & 9.67 & 45.21 & 30.09 & 56.59 & 39.29 \\
  & HuPER & 27.72 & 45.14 & 43.28 & 6.68 & 39.37 & 37.86 & 68.54 & 48.78 \\
  & PhoneticXeus & 25.23 & 47.80 & 42.73 & 9.31 & 41.79 & 33.96 & 60.11 & 43.40 \\
  & BabAR & 16.67 & 31.56 & 28.74 & 6.15 & 27.49 & 49.31 & 75.04 & 59.51 \\
  & Azure Speech & 19.06 & 37.55 & 29.29 & 14.41 & 32.28 & 42.07 & 61.51 & 49.96 \\
  & Gemini-3.8-Flash & 12.74 & 31.03 & 23.67 & 19.86 & 26.86 & 51.78 & 66.61 & 58.26 \\
  & GPT-Audio-1.5 & 26.56 & 51.73 & 40.88 & 15.11 & 41.11 & 32.45 & 58.35 & 41.71 \\
  & Qwen2.5-Omni-7B & 15.90 & 40.88 & 32.62 & 13.36 & 35.41 & 38.37 & 58.88 & 46.46 \\
  & Qwen2.5-Omni-7B + TIMIT \textbf{(ours)} & 12.57 & 29.89 & 25.70 & 8.61 & 26.15 & 50.92 & 72.93 & 59.97 \\
  & Qwen2.5-Omni-7B + TIMIT+SC \textbf{(ours)} & \textbf{10.76} & 26.23 & 22.06 & 6.68 & 22.84 & 54.71 & \textbf{75.57} & 63.47 \\
  & Audio Flamingo 3 & 64.92 & 79.83 & 61.63 & 9.31 & 54.75 & 24.36 & 64.85 & 35.41 \\
  & Audio Flamingo 3$_{\mathrm{TIMIT}}$ \textbf{(ours)} & 13.26 & 31.03 & 27.06 & 6.33 & 26.89 & 49.29 & 73.46 & 59.00 \\
  & Audio Flamingo 3$_{\mathrm{TIMIT+SC}}$ \textbf{(ours)} & 11.48 & \textbf{26.03} & \textbf{21.49} & 8.96 & \textbf{22.69} & \textbf{55.61} & 74.87 & \textbf{63.82} \\
  \midrule
  \multirow{18}{*}{Acted}
  & W2V2P-LV60 & 6.02 & 17.61 & 16.07 & 7.73 & 16.60 & 50.72 & 81.16 & 62.43 \\
  & W2V2P-XLSR53 & 5.84 & 13.93 & 13.51 & 3.96 & 13.21 & 57.69 & 88.02 & 69.70 \\
  & W2V2P-XLSR53$_{\mathrm{SW}}$ \textbf{(ours)} & 5.50 & 12.72 & 12.10 & 4.54 & 12.04 & 60.34 & 88.21 & 71.66 \\
  & MultiIPA & 15.90 & 53.07 & 48.90 & 5.60 & 46.70 & 19.89 & 63.38 & 30.28 \\
  & ZIPA-CTC & 3.13 & 10.42 & 8.60 & 9.18 & 9.86 & 66.87 & 84.83 & 74.79 \\
  & ZIPA-CTC-NS & 4.64 & 12.93 & 10.62 & 10.14 & 12.16 & 60.36 & 81.35 & 69.30 \\
  & HuPER & 5.02 & 12.38 & 12.20 & 2.61 & 11.66 & 60.87 & 90.63 & 72.83 \\
  & PhoneticXeus & 6.96 & 16.71 & 15.77 & 3.96 & 15.79 & 51.85 & 84.15 & 64.16 \\
  & BabAR & 64.39 & 71.50 & 71.13 & \textbf{1.64} & 68.20 & 11.16 & 43.96 & 17.80 \\
  & Azure Speech & 4.14 & 8.59 & 5.63 & 9.28 & 8.11 & 71.20 & 81.45 & 75.98 \\
  & Gemini-3.8-Flash & 3.01 & 10.53 & 6.00 & 20.97 & 10.05 & 70.35 & 73.14 & 71.72 \\
  & GPT-Audio-1.5 & 2.66 & 9.32 & 6.62 & 12.75 & 8.85 & 71.24 & 81.84 & 76.17 \\
  & Qwen2.5-Omni-7B & 8.76 & 24.59 & 20.41 & 15.94 & 22.55 & 39.36 & 68.12 & 49.89 \\
  & Qwen2.5-Omni-7B$_{\mathrm{TIMIT}}$ \textbf{(ours)} & 3.41 & 8.94 & 7.89 & 5.31 & 8.53 & 69.67 & 88.79 & 78.08 \\
  & Qwen2.5-Omni-7B$_{\mathrm{TIMIT+SW}}$ \textbf{(ours)} & \textbf{0.92} & \textbf{3.50} & \textbf{2.61} & 2.90 & \textbf{3.34} & \textbf{86.75} & \textbf{93.62} & \textbf{90.06} \\
  & Audio Flamingo 3 & 20.55 & 27.54 & 25.01 & 5.22 & 23.95 & 38.35 & 80.97 & 52.05 \\
  & Audio Flamingo 3$_{\mathrm{TIMIT}}$ \textbf{(ours)} & 4.35 & 10.00 & 9.15 & 4.15 & 9.53 & 66.38 & 88.89 & 76.00 \\
  & Audio Flamingo 3$_{\mathrm{TIMIT+SW}}$ \textbf{(ours)} & 1.06 & 3.97 & 2.82 & 4.15 & 3.79 & 85.71 & 92.17 & 88.83 \\
  \bottomrule
  \end{tabular}%
  }
  \caption{Benchmarked results on our child and acted speech evaluation datasets.}
  \label{tab:diagnostic-results}
  \end{table*}

\begin{table*}[t]
  \centering
  \scriptsize
  \setlength{\tabcolsep}{3pt}
  \renewcommand{\arraystretch}{1.05}
  \resizebox{\textwidth}{!}{%
  \begin{tabular}{lrrrr|rrrr|rrrr}
  \toprule
  \textbf{Model}
  & \multicolumn{4}{c|}{\textbf{Word}}
  & \multicolumn{4}{c|}{\textbf{Sentence}}
  & \multicolumn{4}{c}{\textbf{Passage}} \\
  \cmidrule(lr){2-5} \cmidrule(lr){6-9} \cmidrule(lr){10-13}
  & \textbf{FRR$\downarrow$} & \textbf{Prec.$\uparrow$}
  & \textbf{Rec.$\uparrow$} & \textbf{F1$\uparrow$}
  & \textbf{FRR$\downarrow$} & \textbf{Prec.$\uparrow$}
  & \textbf{Rec.$\uparrow$} & \textbf{F1$\uparrow$}
  & \textbf{FRR$\downarrow$} & \textbf{Prec.$\uparrow$}
  & \textbf{Rec.$\uparrow$} & \textbf{F1$\uparrow$} \\
  \midrule
  W2V2P-LV60
  & 15.14 & 51.34 & 79.30 & 62.33 & 9.33 & 23.50 & 71.34 & 35.36 & 10.44 & 6.49 & 77.39 & 11.98 \\
  W2V2P-XLSR53
  & 18.97 & 45.64 & 79.94 & 58.10 & 13.87 & 19.09 & 80.57 & 30.87 & 15.08 & 4.82 & 81.53 & 9.10 \\
  MultiIPA
  & 54.53 & 16.62 & 56.69 & 25.70 & 48.86 & 4.85 & 64.33 & 9.01 & 46.83 & 1.21 & 64.33 & 2.38 \\
  ZIPA-CTC
  & 11.16 & 57.85 & 78.66 & 66.67 & 9.67 & 25.45 & 81.53 & 38.79 & 9.44 & \textbf{7.76} & 84.71 & \textbf{14.22} \\
  ZIPA-CTC-NS
  & 9.17 & 61.68 & 77.39 & 68.64 & 9.44 & 25.20 & 78.98 & 38.21 & 9.47 & 7.52 & 82.17 & 13.78 \\
  HuPER
  & 16.00 & 53.64 & \textbf{89.17} & 66.99 & 19.97 & 14.80 & \textbf{86.62} & 25.28 & 21.58 & 3.69 & \textbf{88.85} & 7.09 \\
  PhoneticXeus
  & 12.20 & 56.98 & 81.85 & 67.19 & 11.98 & 21.38 & 80.89 & 33.82 & 12.19 & 5.81 & 80.25 & 10.84 \\
  BabAR
  & 74.30 & 9.56 & 39.17 & 15.37 & 76.45 & 2.17 & 42.04 & 4.13 & 65.11 & 0.86 & 59.87 & 1.69 \\
  Azure Speech
  & \textbf{5.60} & \textbf{70.69} & 78.34 & \textbf{74.32} & 8.91 & \textbf{26.10} & 79.30 & \textbf{39.27} & 9.86 & 7.10 & 80.57 & 13.06 \\
  Gemini-3.8-Flash
  & 8.36 & 62.67 & 71.66 & 66.86 & 9.13 & 9.65 & 23.89 & 13.75 & 9.47 & 1.53 & 15.61 & 2.79 \\
  GPT-Audio-1.5
  & 14.24 & 49.67 & 71.97 & 58.78 & \textbf{8.63} & 21.54 & 57.96 & 31.41 & \textbf{9.12} & 4.85 & 49.36 & 8.82 \\
  Qwen2.5-Omni-7B
  & 21.39 & 36.70 & 65.92 & 47.15 & 19.48 & 12.75 & 72.29 & 21.67 & 17.95 & 3.41 & 69.75 & 6.50 \\
  Qwen2.5-Omni-7B$_{\mathrm{TIMIT}}$
  & 10.05 & 63.62 & 86.31 & 73.24 & 17.22 & 16.96 & \textbf{86.62} & 28.36 & 19.01 & 4.21 & \textbf{88.85} & 8.04 \\
  Audio Flamingo 3
  & 25.24 & 37.39 & 79.30 & 50.82 & 65.47 & 2.87 & 48.73 & 5.43 & 92.96 & 0.26 & 25.80 & 0.51 \\
  Audio Flamingo 3$_{\mathrm{TIMIT}}$
  & 13.73 & 56.69 & 86.31 & 68.43 & 20.16 & 14.43 & 83.76 & 24.61 & 36.60 & 2.06 & 82.17 & 4.01 \\
  \bottomrule
  \end{tabular}%
  }
  \caption{Benchmarked results on our synthetic speech evaluation dataset.}
  \label{tab:synthetic-full-results}
  \end{table*}

\begin{promptbox}
\small
\textbf{Insight 1:} SOTA speech models cannot yet reliably detect phonetic errors on L2-ARCTIC and CMU-Kids, and the evidence suggests potential inconsistencies in the ground-truth phoneme transcriptions.
\end{promptbox}



\subsection{Can SOTA speech models detect phonetic errors in our novel PED datasets?}

As shown in Figure~\ref{tab:diagnostic-results}, the best-performing PR model (BabAR: 59.51 F1), specialized PED model (Azure: 49.96 F1), and SLM (Gemini: 58.26 F1) still fail to reliably detect phonetic errors in our word-level child-speech dataset, despite its transcriptions being agreed upon by two independent annotators. For our acted dataset, which is more controlled and designed to reduce ambiguity in phoneme realization, the best-performing PR model (ZIPA-CTC: 74.79 F1), specialized PED model (Azure: 75.98 F1), and SLM (GPT-Audio: 76.17 F1) also fail to perform PED reliably. Their low precision suggests frequent false rejections.

As shown in Figure~\ref{tab:synthetic-full-results}, even on our synthetic evaluation data, where speech is generated from specified phoneme sequences to produce objectively defined phonetic errors, the best-performing models at the word (Azure: 74.32 F1), sentence (Azure: 39.27 F1), and passage (ZIPA-CTC: 14.22 F1) levels fail to perform PED reliably. Performance degrades substantially as audio length increases, while recall remains stable across levels, suggesting that the degradation is primarily driven by lower precision and more false rejections.

Results reported in Tables~\ref{tab:public-results},~\ref{tab:diagnostic-results}, and \ref{tab:synthetic-full-results} are all unscripted results to fairly compare between models. Specialized PED and SLMs can also be evaluated in the scripted setting where canonical phoneme sequence can also be used as the model input. As shown in Figure~\ref{fig:script}, the results are mixed, and canonical scripting did not consistently help improve models; more detailed discussion can be found in Appendix~\ref{sec:script}.

\begin{promptbox}
\small
\textbf{Insight 2:} SOTA speech models also struggle to detect phonetic errors, even under controlled conditions with acted and synthetic speech, where phoneme realizations are more objectively defined with reduced acoustic variability.
\end{promptbox}


\subsection{Can synthetic data help fine-tune speech models for better PED?}

As shown in Table~\ref{tab:diagnostic-results}, our LoRA fine-tuned SLMs significantly outperformed other model candidates. In our acted dataset, the fine-tuned Qwen achieved 90.06 F1 and the fine-tuned AF3 achieved 88.83 F1 , which is better than the best performing SLM model (GPT-Audio: 76.17 F1), specialized PED (Azure: 75.98 F1) and PR model (ZIPA-CTC: 74.79 F1). In our child dataset, the fine-tuned AF3 achieved 63.82 F1 and the fine-tuned Qwen achieved 63.47 F1, outperforming the best performing SLM model (Gemini: 58.26), specialized PED (Azure: 49.96 F1) and PR model (BabAR: 59.51 F1). Although the fine-tuned Qwen and AF3 models performed strongly on acted speech, their performance on child speech remains unsatisfactory. This may reflect the distribution shift between the adult speech used for training (TIMIT + SC) and the child speech used for evaluation. These results demonstrate the promise of using synthetic data to scale quantity and diversity of training data. However, the sim-to-real and adult-to-child gap may still hinder model generalization.

\begin{promptbox}
\small
\textbf{Insight 3:} Synthetic speech containing common phonetic errors offers a promising way to scale PED training data, but current phoneme-controlled TTS systems need further improvements to narrow the sim-to-real and adult-to-child gaps.
\end{promptbox}

\subsection{Error Analysis}

\begin{figure}
    \centering
    \includegraphics[width=\linewidth]{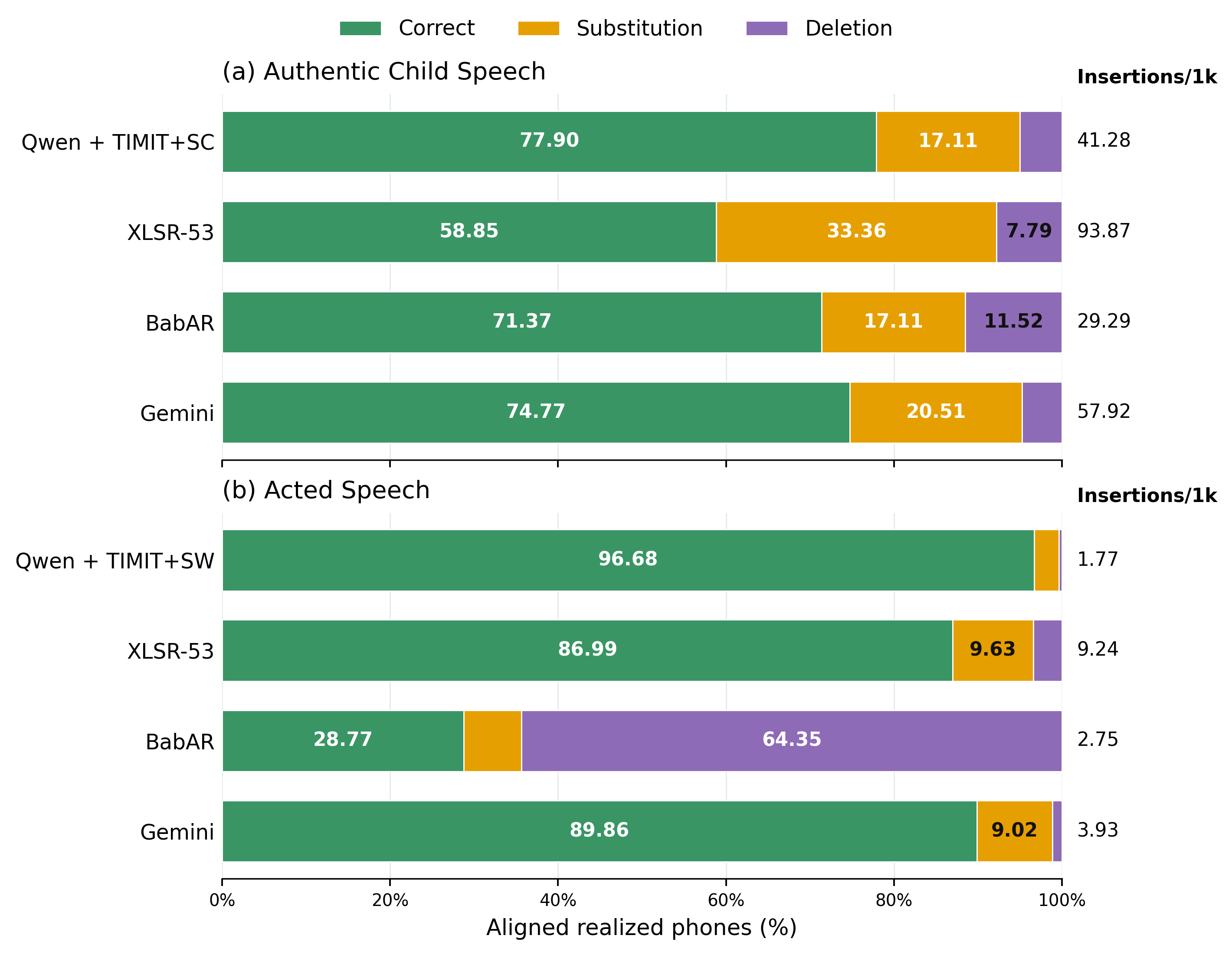}
    \caption{Error profiles for four representative models on Authentic Child Speech and Acted Speech. }
    \label{fig:error-profiles}
\end{figure}

To identify failure modes obscured by aggregate metrics, we analyze four representative models: Qwen2.5-Omni-7B fine-tuned on TIMIT and synthetic data, W2V2P-XLSR53 for specialized phoneme recognition, BabAR for child-speech models, and Gemini for commercial SLMs. We categorize each aligned realized phone as correctly predicted, substituted, or deleted, and separately report insertions per 1,000 realized phones (Figure~\ref{fig:error-profiles}). On acted speech, Qwen fine-tuned on TIMIT and SW data achieves the lowest substitution, deletion, and insertion rates among these four models. BabAR, pretrained and fine-tuned exclusively on child speech, performs substantially worse on acted adult speech but remains competitive on our child-speech dataset. Phone-level confusion matrices and insertion inventories for all four models are provided in Appendix~\ref{sec:appendix-error-analysis}. The models struggle to distinguish acoustically similar sounds, including /θ/–/f/, /ʃ/–/t͡ʃ/, /ɔ/–/ɑ/, and /eɪ/–/ɪ/. These confusions motivate targeted synthetic training data to help models learn the phonetic boundaries between these sounds. The difficulty of distinguishing /θ/ from /f/ using audio alone also motivates future work on audio-visual speech recognition. We hypothesize that tongue and mouth movements may provide noise-invariant cues for robust PED.

\section{Conclusion}

In this work, we introduced PEDBench, the first standardized benchmark for evaluating SOTA speech models on phonetic error detection. We found that all SOTA speech models still fail to reliably detect phonetic errors across word- and sentence-level speech in both real-world and highly controlled settings. We explored synthetic data for fine-tuning SLMs and demonstrated their potential to scale PED training data, especially for rare errors that are difficult to collect in the real world. This work lays the groundwork for more accurate evaluation and effective training of PED models toward reliable deployment in real-world applications.

\section{Limitations}

While we believe PEDBench can support the future evaluation and development of PED systems, it has the following limitations:

\paragraph{Evaluation Goals.} PEDBench aims to expose the limitations of SOTA speech models through diagnostic datasets with objectively defined phonetic errors. However, its three new datasets do not yet provide a comprehensive evaluation of PED. Future work should collect real-world speech covering a broader range of phonetic errors from speakers with diverse linguistic backgrounds and speech and language abilities. These expanded datasets would support more comprehensive model validation before real-world deployment.

\paragraph{Fine-Tuning Claims.} Our results demonstrate the potential of synthetic data generated through phoneme-controlled TTS to scale training datasets. However, they do not yet establish that our fine-tuned models can robustly detect phonetic errors. The current pipeline remains constrained by the ability of TTS models to reliably produce the intended phonetic errors and requires human review to exclude inaccurately generated samples. Future work should develop TTS models that generate these errors more accurately and provide finer control over speech production, acoustic variability, and speaker identity.

\paragraph{Phoneme Transcription.} Despite our best efforts to ensure annotation quality through independent transcription or verification by two phonetically trained annotators, some ambiguity and subjectivity may remain. Our acted and synthetic speech datasets help reduce this ambiguity by introducing phonetic errors under controlled conditions, but they may not eliminate it entirely.

\paragraph{Modality.} The models evaluated in this work use only audio input; we have not yet explored audio-visual input. Incorporating visual information may help models distinguish acoustically similar phonemes, such as /θ/ and /f/, by capturing differences in visible articulatory movements.

\paragraph{Language.} PEDBench focuses on phonetic error detection in English and uses the CMU-39 phoneme inventory to standardize evaluation. All three new PEDBench datasets are in English, limiting the benchmark's applicability to multilingual settings.

\paragraph{Benchmarked Models.} PEDBench evaluates prominent state-of-the-art models across three categories: specialized phoneme recognizers, specialized PED models, and speech-language models. However, the evaluation is not exhaustive, and some available models are not represented in this work.

\section{Ethical Considerations}

For the data, we downloaded publicly available datasets and complied with their respective licenses. In particular, because the CMU Kids license prohibits transmitting the data to third parties, we did not evaluate commercial models on CMU Kids through their APIs. For our new PED datasets, our university's Institutional Review Board (IRB) approved the data collection protocol and data security plan. The protocol explicitly included consent language permitting the use of the collected data to evaluate third-party commercial automatic speech recognition (ASR) models. However, the consent did not permit public release of the child speech data. Accordingly, only our acted and synthetic datasets can be released publicly. We plan to retain the child speech dataset as a private held-out evaluation set for a future PED leaderboard.

For the models, we benchmarked existing speech models and fine-tuned open-source speech-language models. Upon acceptance of the paper, we will release our fine-tuned model checkpoints in accordance with the licenses of the underlying models. Our code builds on the PRiSM toolkit \cite{bharadwaj2026prism}, which in turn follows the Lightning-Hydra-Template (MIT license). We will release our code under a license compatible with both and retain the original copyright notices.
\section{Generative AI Disclaimer}
In this paper, AI coding agents were used to assist with development, and the authors verified all resulting code. AI writing assistants were also used to proofread the manuscript, correct grammatical errors, and improve its overall language. The authors reviewed and verified all AI-assisted text.

\bibliography{custom}

\appendix

\section{Additional Benchmark Details}

\subsection{Public Datasets}
\label{sec:public-data}

\paragraph{L2-ARCTIC-Perceived.}
L2-ARCTIC-Perceived is one of the most commonly used public benchmark in the PED and MDD literature. It contains sentence reading recordings from 24 non-native adult English speakers, with a total of 3599 utterances~\citep{zhao2018l2}. We used the processed version released by \citet{bharadwaj2026prism}.

\paragraph{CMU Kids.} CMU Kids Corpus is one of the commonly used child speech recognition datasets also with phoneme-level transcription. The corpus comprises 5,180 prompted sentence reading utterances in English from 76 children aged 6--11 years~\citep{Mostow-1996-14274}. We used the version distributed by the Linguistic Data Consortium (LDC).\footnote{\url{https://catalog.ldc.upenn.edu/LDC97S63}}

\subsection{Spectrum of Ambiguity in Phoneme Production}
\label{sec:spectrum}

\begin{figure}[!ht]
    \centering
    \includegraphics[width=\linewidth]{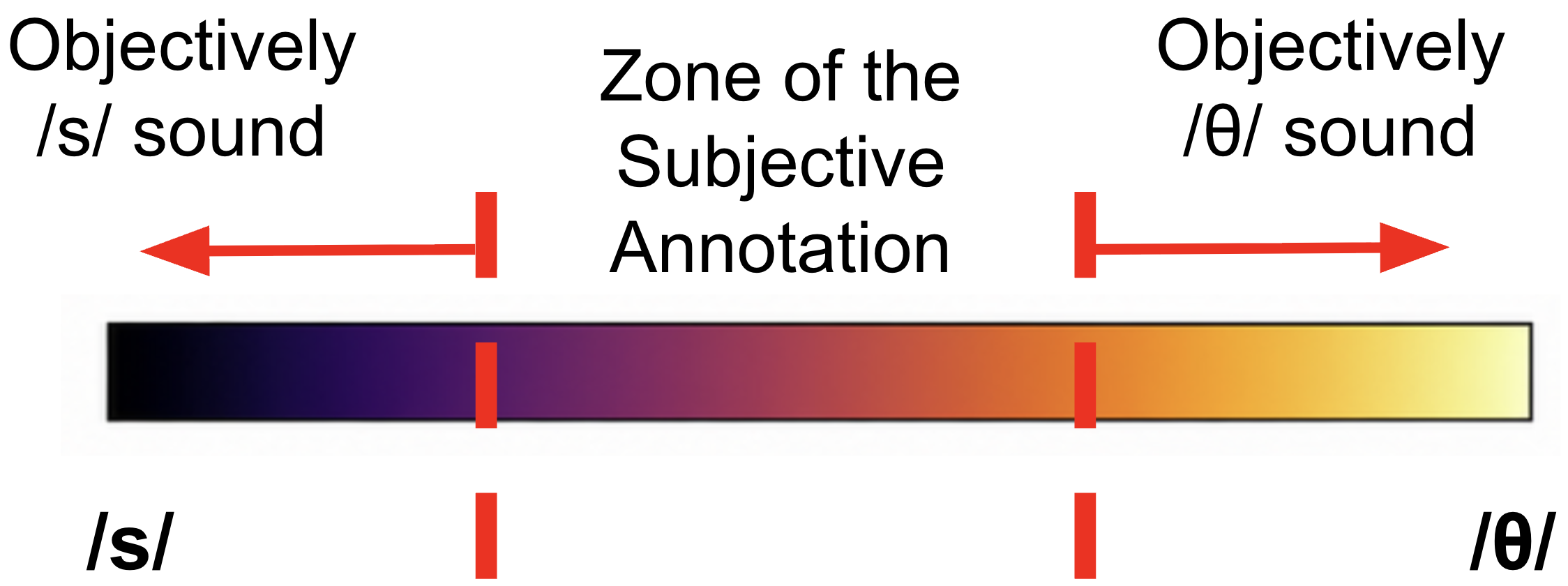}
    \caption{Spectrum of Ambiguity in Phoneme Production }
    \label{fig:spectrum}
\end{figure}

As shown in Figure~\ref{fig:spectrum}, natural phoneme production can be ambiguous, making it difficult for annotators to agree on phoneme transcriptions. For example, /s/ and /θ/ are distinct phonemes, but natural speech, especially child speech, may include realizations near the phonetic boundary between them. This region is highlighted as the zone of subjective annotation, where different annotators may assign different transcriptions. Such ambiguity makes it inherently challenging to establish consistent ground-truth transcriptions for real-world PED datasets.

\section{IPA Substitution Pairs}
\label{sec:ipa-substitution-pairs}

Table~\ref{tab:ipa-substitution-pairs} lists the substitution pairs used to generate the synthetic mispronunciations described in Section~\ref{subsubsec:novel-diagnostic-data}.

\begin{table}[t]
\centering
\footnotesize
\setlength{\tabcolsep}{2.4pt}
\renewcommand{\arraystretch}{1.1}
\begin{tabular}{@{}ccc@{\hspace{10pt}}cccc@{}}
\toprule
\multicolumn{3}{c}{\textbf{Vowel}} &
\multicolumn{4}{c}{\textbf{Consonant}} \\
\cmidrule(r){1-3}\cmidrule(l){4-7}
(ɪ, i)   & (ɪ, ɛ)   & (i, ɛ)    & (b, p)   & (b, v)   & (tʃ, t)  & (tʃ, ʃ) \\
(i, eɪ)  & (ɛ, æ)   & (ɛ, ɪ)    & (d, k)   & (d, g)   & (f, p)   & (f, b)  \\
(eɪ, ɛ)  & (eɪ, æ)  & (æ, ɑ)    & (g, d)   & (g, k)   & (dʒ, d)  & (dʒ, z) \\
(æ, ɛ)   & (ɑ, oʊ)  & (ɑ, u)    & (k, t)   & (k, g)   & (l, w)   & (l, j)  \\
(ɔ, oʊ)  & (ɔ, u)   & (oʊ, ɔ)   & (m, b)   & (m, n)   & (n, m)   & (n, t)  \\
(oʊ, aʊ) & (ʊ, u)   & (ʊ, oʊ)   & (ŋ, n)   & (ŋ, k)   & (p, b)   & (p, f)  \\
(u, ʊ)   & (u, oʊ)  & (ɝ, ɛ)    & (r, w)   & (r, l)   & (s, θ)   & (s, d)  \\
(ɝ, ʊ)   & (aɪ, ɑ)  & (aɪ, eɪ)  & (ʃ, s)   & (ʃ, t)   & (t, k)   & (t, d)  \\
(ɔɪ, ɔ)  & (ɔɪ, aɪ) & (ə, oʊ)   & (ð, d)   & (ð, v)   & (θ, f)   & (θ, t)  \\
(ə, ɪ)   &          &           & (v, b)   & (v, f)   & (w, r)   & (w, v)  \\
         &          &           & (j, w)   & (j, ʒ)   & (z, θ)   & (z, d)  \\
\bottomrule
\end{tabular}
\caption{IPA substitution pairs, written as (canonical, substituted), used to generate mispronunciations in the synthetic data.}
\label{tab:ipa-substitution-pairs}
\end{table}

\section{Evaluation Metrics}
\label{sec:appendix-metrics}

\paragraph{Phoneme Recognition Metrics.}
Recent PED work commonly reports phoneme-recognition metrics. Let $d$ denote the unit-cost Levenshtein
distance and $|U_i^{39}|$ the number of reference phonemes in utterance
$i$. We report the corpus-level phoneme error rate
$\mathrm{PER}=\sum_i d\bigl(U_i^{39},\hat{U}_i^{39}\bigr)\,/\,\sum_i
|U_i^{39}|$ between the uttered and predicted sequences, and, following
\citet{bharadwaj2026prism}, the phonetic feature error rate
$\mathrm{PFER}=\sum_i d_{\phi}\bigl(U_i^{39},\hat{U}_i^{39}\bigr)\,/\,
\sum_i |U_i^{39}|$, where $d_{\phi}$ is an edit distance whose
substitution cost is the articulatory-feature distance between phonemes.
PFER therefore penalizes confusions between articulatory similar
phonemes less than dissimilar ones.

\paragraph{PED Metrics.}

PER and PFER aggregate performance over all phone positions, whereas phonetic errors are sparse; consequently, a model may perform well on PER/PFER metrics while missing the relatively few errors that matter. We therefore adopt the hierarchical evaluation framework used in the mispronunciation detection and diagnosis (MDD) literature~\citep{li2016mispronunciation,daoud2025phoneme}. 
\begin{gather*}
  \mathrm{FRR}=\frac{\mathrm{FR}}{\mathrm{TA}+\mathrm{FR}},
  \qquad
  \mathrm{FAR}=\frac{\mathrm{FA}}{\mathrm{TR}+\mathrm{FA}},\\
  \mathrm{DER}=
  \frac{\mathrm{FR}+\mathrm{FA}+\mathrm{DE}}
  {\mathrm{TA}+\mathrm{FR}+\mathrm{TR}+\mathrm{FA}},\\
  \mathrm{Precision}_{\mathrm{DW}}=
  \frac{\mathrm{CD}}{\mathrm{TR}+\mathrm{FR}},\\
  \mathrm{Recall}_{\mathrm{DW}}=
  \frac{\mathrm{CD}}{\mathrm{TR}+\mathrm{FA}},\\
  \mathrm{F1}_{\mathrm{DW}}=
  \frac{2\,\mathrm{CD}}
  {2\,\mathrm{TR}+\mathrm{FR}+\mathrm{FA}}.
  \end{gather*}

FRR measures how often correctly pronounced phones are incorrectly flagged, a failure mode that can undermine user trust in deployment. FAR measures how often actual phonetic errors are incorrectly accepted as canonical. DER captures all incorrect detection and diagnosis outcomes: false rejections, false acceptances, and correctly detected errors assigned the wrong realized phone.

\section{MDD Alignment and Metrics}
\label{sec:mdd-alignment}

We followed the standard MDD literature to align the canonical, uttered and predicted phoneme sequences~\cite{gelin2023comparing}.

For each utterance we align the canonical sequence $C$, the perceived sequence $U$, and the model prediction $P$. All three are first mapped to CMU-39 (see Appendix~\ref{sec:cmu39}). 

First, we align $C$ to $U$ by minimum edit distance without looking at $P$. This fixes the mispronunciation events, so every model is scored against the same events. If two alignments have the same cost, we select the one with fewer rows and break remaining ties by preferring a match or substitution, then a deletion, then an insertion.

Second, we place $P$ onto these fixed events. The alignment first minimizes disagreements with $U$. Among equally good alignments, it prefers more exact matches with $U$, then fewer extra rows, then more matches with $C$. A predicted phone that fits no event becomes its own insertion row.

Third, we label each aligned row $(c, u, p)$. When the speaker said the phone correctly ($c = u$), the row is a true acceptance (TA) if the model also outputs $c$, and a false rejection (FR) if it outputs anything else. When the speaker made an error ($c \neq u$), the row is a false acceptance (FA) if the model outputs the canonical phone $c$, a correct diagnosis (CD) if it outputs what the speaker said ($u$, or nothing if the speaker dropped the phone), and a diagnosis error (DE) otherwise. True rejections are TR $=$ CD $+$ DE. A phone the model inserts where nothing was said, row $(\varepsilon, \varepsilon, p)$, counts as FR. We sum these counts over all rows of all utterances.

From these counts we compute the false rejection rate, false acceptance rate, overall diagnostic error rate, and diagnosis-weighted precision, recall, and F1 defined in Section~\ref{subsec:metrics}. Standard MDD precision and recall put TR in the numerator; the diagnosis-weighted versions use CD, so a model gets credit only when it also names the phone the speaker produced. PER is the edit distance between $P$ and $U$ divided by the length of $U$.

\section{Phone Normalization and CMU-39 Projection}
\label{sec:cmu39}

\begin{table}[t]
\centering
\footnotesize
\setlength{\tabcolsep}{6pt}
\renewcommand{\arraystretch}{1.1}
\begin{tabular}{@{}ll@{}}
\toprule
\textbf{Source symbols} & \textbf{CMU-39} \\
\midrule
ə \textturna\ \textreve\ \textrevepsilon\ \textcloserevepsilon\ \textramshorns & ʌ \\
\textbari & ɪ \\
\textrhookschwa\ ɝ \textipa{3\textrhoticity} & \textipa{@\textrhoticity} \\
e / o & eɪ / oʊ \\
a \textturnscripta & ɑ \\
œ / ø \textbaro\ \textscy\ / y \textbaru\ \textturnm & ɛ / ʊ / u \\
\textltilde\ \textbeltl\ \textrtaill\ \textscl; \textturny & l; l j \\
\textltailm\ / \textrtailn\ \textscn\ / \textltailn & m / n / n j \\
\textturnrrtail\ \textinvscr\ \textscr\ r & \textturnr \\
x \textchi\ \textcrh\ \texthth\ \textrevglotstop & h \\
\textphi\ / \textscriptv & f / v \\
\textctc\ \textrtails\ ç / \textctz\ \textrtailz & ʃ / ʒ \\
c q / \textbardotlessj\ \textscg\ \textgamma & k / ɡ \\
\textrtailt\ / \textrtaild\ \texthtd & t / d \\
tʃ \textteshlig\ t\textctc\ / dʒ \textdyoghlig\ d\textctz & \textipa{\t{tS}} / \textipa{\t{dZ}} \\
ts / dz & t s / d z \\
\textfishhookr\ \textrtailr\ / \textipa{\~R} & t, d / t, d, n \\
\textglotstop\ / \textbeta & t or none / b, v \\
\bottomrule
\end{tabular}
\caption{Main rules for mapping IPA to the CMU-39 inventory. Rhotic hooks on other vowels are removed. The last two rows depend on the aligned canonical phone. Only aɪ, aʊ, eɪ, oʊ, and ɔɪ are single diphthongs; other vowel sequences are split (e.g., əʊ becomes ʌ ʊ).}
\label{tab:cmu39-rules}
\end{table}

\begin{table}[t]
\centering
\footnotesize
\setlength{\tabcolsep}{2pt}
\renewcommand{\arraystretch}{1.1}
\begin{tabular}{@{}llcl@{}}
\toprule
\textbf{Model} & \textbf{Type} & \textbf{Multi.} & \textbf{Native output} \\
\midrule
W2V2P-XLSR53 & CTC & \checkmark & IPA (eSpeak) \\
W2V2P-LV60 & CTC & \checkmark & IPA (eSpeak) \\
MultiIPA & CTC & \checkmark & IPA \\
ZIPA-CTC / -NS & CTC & \checkmark & IPA \\
PhoneticXeus & CTC & \checkmark & IPA \\
HuPER & CTC & $\times$ & ARPAbet (to IPA) \\
BabAR & CTC & \checkmark & IPA \\
Azure Speech & API & $\times$ & IPA (en-US) \\
Gemini-3.8-Flash & SLM, API & \checkmark & CMU-39 (prompted) \\
GPT-Audio-1.5 & SLM, API & \checkmark & CMU-39 (prompted) \\
Qwen2.5-Omni-7B & SLM & \checkmark & CMU-39 (prompted) \\
Audio Flamingo 3 & SLM & -- & CMU-39 (prompted) \\
\bottomrule
\end{tabular}
\caption{Models and their native output. Multi.\ marks models trained on more than one language. W2V2P-LV60 is pretrained on English audio and fine-tuned on multilingual phone labels. Audio Flamingo 3's language coverage is not documented. All outputs are projected to CMU-39.}
\label{tab:model-inventory}
\end{table}

Different models output different alphabets, including multilingual IPA, English IPA, or ARPAbet. To fairly compare them, we map every output to the same 39-phone inventory of 24 consonants and 15 vowels (see Table~\ref{tab:cmu39-rules}). The projection rules were reviewed by reviewers with linguistic and phonetic trainings. This inventory follows CMU-39 but writes each phone in IPA, and merges AH/AX into ʌ and ER/AXR into \textipa{@\textrhoticity}.

The mapping has two steps. Normalization decomposes Unicode, writes affricates with a tie bar, maps variant symbols to standard ones, and removes length, stress, and aspiration marks. Projection then removes spaces and punctuation, and re-reads the string by longest match, mapping each symbol to one of the 39 phones. Because spaces are removed, adjacent symbols can merge into one phone (e.g., a i becomes aɪ). Most diacritics are ignored and the base phone is kept. Symbols with no mapping are removed and logged, not counted as errors themselves.

Six symbols depend on their context. Flaps (\textfishhookr, \textrtailr) become t or d, the nasal flap \textipa{\~R} becomes t, d, or n, the glottal stop \textglotstop\ becomes t, and \textbeta\ becomes b or v. For these, we align the sequence to the projected canonical sequence and use the canonical phone at that position if it is one of the allowed choices. Otherwise, we use a default (t, n, v; the glottal stop is dropped).

Table~\ref{tab:model-inventory} lists each model's native output. Speech-language models are given the CMU-39 inventory in the prompt, but their outputs go through the same projection as all others.

\section{Fine-Tuning Details}
\label{sec:ALMSpec}

We fine-tune two speech-language models (Qwen2.5-Omni-7B and Audio Flamingo 3) with LoRA, and one CTC phone recognizer (W2V2P-XLSR53) with full fine-tuning. Table~\ref{tab:ft-settings} lists the settings. All runs use a single seed (42).

\begin{table}[t]
\centering
\footnotesize
\setlength{\tabcolsep}{3pt}
\renewcommand{\arraystretch}{1.1}
\begin{tabular}{@{}lll@{}}
\toprule
 & \textbf{Qwen / AF3} & \textbf{W2V2P-XLSR53} \\
\midrule
Method & LoRA & Full fine-tuning \\
Trained weights & Text-LM q,k,v,o & All but CNN \\
Trainable params & 10.1M & -- \\
LoRA $r$ / $\alpha$ / dropout & 16 / 32 / 0.05 & -- \\
Loss & Cross-entropy & CTC \\
Optimizer & AdamW, wd 0.01 & AdamW, wd 0.01 \\
Learning rate & $2\times10^{-4}$ & $1\times10^{-5}$ \\
Linear warmup & 3\% & 10\% \\
Effective batch & 16 (1 $\times$ 16) & 16 (4 $\times$ 4) \\
Precision, clip & bf16, 1.0 & bf16, 1.0 \\
Model selection & Best dev PER & 5-fold CV \\
\bottomrule
\end{tabular}
\caption{Fine-tuning settings.}
\label{tab:ft-settings}
\end{table}

\begin{figure*}[t]
\centering
\begin{tcolorbox}[colback=gray!4, colframe=gray!55, boxrule=0.4pt, arc=2pt,
  boxsep=0pt, left=4pt, right=4pt, top=3pt, bottom=3pt]
\scriptsize
\begin{verbatim}
SYSTEM PROMPT
### You are an IPA transcription system.

### Transcribe all and only the speech in the attached audio into the
International Phonetic Alphabet (IPA).

### Allowed IPA Phone Inventory
Use only the following CMU39-style General American IPA phone symbols in the transcription.

Consonants:
p b t d k ɡ t͡ʃ d͡ʒ f v θ ð s z ʃ ʒ h m n ŋ l ɹ w j

Vowels:
i ɪ eɪ ɛ æ ɑ ɔ oʊ ʊ u ʌ ə˞ aɪ aʊ ɔɪ

### Return exactly one minified JSON object in this form:
{"transcription":"<IPA transcription>"}

### Stopping Rule:
Please stop the reponse after {"transcription":"<IPA transcription>"} and end after the } character.

### Rules:
- Output one JSON object and nothing else.
- The transcription value must contain only IPA symbols and spaces.
- Do not explain your answer or provide examples.
- Do not continue the conversation.
- Do not use Markdown, phonemic slashes, or phonetic brackets.
- Stop immediately after the closing brace.

USER PROMPT (UNSCRIPTED)
Transcribe the attached audio into IPA only using IPAs in the allowed IPA Phone Inventory.
Return only {"transcription":"<IPA transcription>"}.

USER PROMPT (SCRIPTED) = unscripted user prompt, followed by:
The canonical reference text and IPA sequence below describe what the speaker was intended to say.
The actual pronunciation may contain phoneme substitutions, deletions, or insertions.

Transcribe the phonemes actually spoken in the audio, including any departures from the intended
pronunciation. Use the canonical reference as supporting context to resolve acoustic ambiguity, not as
an answer to copy. Do not correct the speaker's pronunciation to match the reference, and do not
introduce deviations that are unsupported by the audio.

Return only the required transcription JSON object; do not provide explanations or separate error labels.

Canonical reference text: <word>
Canonical phoneme sequence (IPA): <canonical IPA>
\end{verbatim}
\end{tcolorbox}
\caption{System prompt and user prompts for speech-language models. Long lines are wrapped for display.}
\label{fig:ALMSpec-prompt}
\end{figure*}

\subsection{Speech-Language Models}
\label{sec:ALMSpec-alm}

We add LoRA adapters to the query, key, value, and output projections of every self-attention layer in the text language model for Qwen2.5-Omni-7B and Audio Flamingo 3. This gives 112 adapted modules and 10.1M trainable parameters for each model. The audio encoder and all other weights stay frozen. The model is asked for a space-separated sequence of CMU-39 IPA phones as a value in a JSON object. We compute loss only on the answer tokens. We train on one GPU in bf16 with AdamW (learning rate $2\times10^{-4}$, weight decay 0.01), a linear schedule with 3\% warmup, an effective batch of 16 utterances, and gradient clipping at 1.0. After each epoch, we decode the whole development set greedily and keep the checkpoint with the lowest PER. We train each model for 6 epochs.

\begin{table*}[t]
\centering
\footnotesize
\setlength{\tabcolsep}{5pt}
\renewcommand{\arraystretch}{1.1}
\begin{tabular}{@{}lrrrll@{}}
\toprule
\textbf{Data} & \textbf{Train} & \textbf{Dev.} & \textbf{Test} & \textbf{Split} & \textbf{Used in} \\
\midrule
TIMIT & 3,696 & 400 & 192 & Kaldi speaker split, no SA & TIMIT, TIMIT+SW, TIMIT+SC (train) \\
Synthetic words (SW) & 396 & -- & -- & All for training & TIMIT+SW, TIMIT+SC \\
Synthetic combined (SC) & 550 & -- & -- & All for training & TIMIT+SC \\
Acted & -- & 1,188 & -- & All for model selection & TIMIT+SC \\
\midrule
TIMIT+SW & 4,092 & 400 & -- & Dev.\ = TIMIT & Evaluated on Acted \\
TIMIT+SC & 4,246 & 1,188 & -- & Dev.\ = Acted & Evaluated on Authentic \\
\bottomrule
\end{tabular}
\caption{Data used for fine-tuning and model selection.}
\label{tab:ALMSpec-data-split-summary}
\end{table*}

\subsection{Scripted Training}
\label{sec:ALMSpec-scripted}

For scripted runs, the user prompt also contains the intended word and its canonical phone sequence (Figure~\ref{fig:ALMSpec-prompt}). The same prompt is used for training and testing. The model only sees the perceived transcription as the training target. The canonical sequences for TIMIT utterances come from the TIMIT pronunciation dictionary.


\subsection{Training Data}
\label{sec:ALMSpec-data-splits}

Table~\ref{tab:ALMSpec-data-split-summary} lists the data used for training and model selection. TIMIT follows the standard Kaldi speaker split. We drop the repeated SA sentences, which leaves 3,696 training utterances, 400 development utterances from 50 speakers, and the 192-utterance core test set from 24 speakers. SW is the 396 isolated-word recordings of our synthetic set. SC adds 154 synthetic recordings of error patterns from authentic child speech, for 550 in total. In both SW and SC, all synthetic recordings go to training.

We choose the best checkpoint for TIMIT and TIMIT+SW models using the 400 TIMIT development utterances, and for TIMIT+SC models using the 1,188 Acted utterances. Because Acted utterances were used during checkpoint selection, we do not run any TIMIT+SC models on the Acted benchmark. Also, because the SW recordings are part of the synthetic test set, we do not evaluate TIMIT+SW or TIMIT+SC models on Synthetic. We evaluate TIMIT-only models on all five test sets, TIMIT+SW models on Acted, and TIMIT+SC models on Authentic. Since the words in SW and SC are seen in Authentic and Acted data (but not the audio samples), the Acted and Authentic results for these mixes measure transfer from synthetic to human speech on seen items, not generalization to new items.

\subsection{W2V2P-XLSR53 CTC Fine-Tuning}
\label{sec:ALMSpec-xlsr}

We fine-tune the released checkpoint \nolinkurl{facebook/wav2vec2-xlsr-53-espeak-cv-ft} with CTC loss. We keep the model's own IPA output vocabulary and output layer, and respell three CMU-39 phones to match it: the affricates are written without a tie bar, and the r-colored schwa is written as the single symbol \textrhookschwa. All weights except the convolutional feature encoder are trained. We use a learning rate of $1\times10^{-5}$, a weight decay of 0.01, a linear schedule with 10\% warmup, and bf16.

We train one W2V2P-XLSR53 + SW model on the 396 SW recordings (evaluated on Acted) and one W2V2P-XLSR53 + SC model on the 550 SC recordings (evaluated on Authentic). Because all synthetic recordings are used for training, we pick the number of training steps by five-fold cross-validation grouped by source recording, taking the earliest step whose held-out PER is within one standard error of the best, and then retrain on all recordings for that many steps (25 for SW, 44 for SC).

\begin{table}[t]
\centering
\footnotesize
\setlength{\tabcolsep}{4pt}
\renewcommand{\arraystretch}{1.15}
\begin{tabular}{@{}p{3.9cm}p{3.4cm}@{}}
\toprule
\textbf{Output} & \textbf{Result} \tabularnewline
\midrule
\raggedright\ttfamily \{"transcription":"ʃ i t"\} & \raggedright Accepted. \tabularnewline
\raggedright\ttfamily \{"answer":"ʃ i t"\} & \raggedright Accepted: only one key. \tabularnewline
\raggedright\ttfamily ʃ i t & \raggedright Accepted: plain phone string. \tabularnewline
\raggedright\ttfamily The transcription is \{"transcription":"ʃ i t"\} & \raggedright Invalid: text before the JSON. \tabularnewline
\raggedright\ttfamily The word is sheet. & \raggedright Invalid: English text. \tabularnewline
\bottomrule
\end{tabular}
\caption{Examples of model outputs and how they are parsed. Invalid outputs are scored as empty predictions.}
\label{tab:ALMSpec-output-validation}
\end{table}

\begin{figure*}[t]
    \centering
    \includegraphics[width=\textwidth]{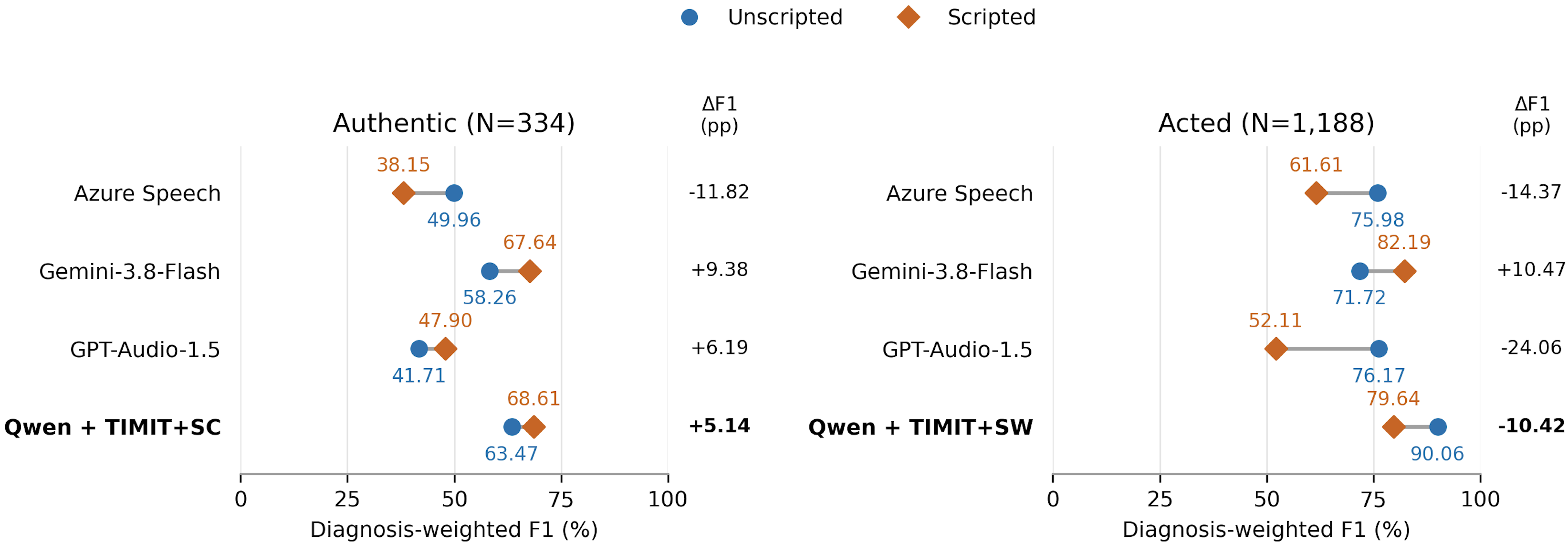}
    \caption{Scripted vs Unscripted Results on Our Authentic and Acted Evaluation Datasets. }
    \label{fig:script}
\end{figure*}

 \begin{table*}[t]
  \centering
  \scriptsize
  \setlength{\tabcolsep}{3pt}
  \renewcommand{\arraystretch}{1.05}
  \resizebox{\textwidth}{!}{%
  \begin{tabular}{lrrrr|rrrr|rrrr}
  \toprule
  \textbf{Model}
  & \multicolumn{4}{c|}{\textbf{Word}}
  & \multicolumn{4}{c|}{\textbf{Sentence}}
  & \multicolumn{4}{c}{\textbf{Passage}} \\
  \cmidrule(lr){2-5}
  \cmidrule(lr){6-9}
  \cmidrule(lr){10-13}
  & \textbf{FRR$\downarrow$}
  & \textbf{Prec.$\uparrow$}
  & \textbf{Rec.$\uparrow$}
  & \textbf{F1$\uparrow$}
  & \textbf{FRR$\downarrow$}
  & \textbf{Prec.$\uparrow$}
  & \textbf{Rec.$\uparrow$}
  & \textbf{F1$\uparrow$}
  & \textbf{FRR$\downarrow$}
  & \textbf{Prec.$\uparrow$}
  & \textbf{Rec.$\uparrow$}
  & \textbf{F1$\uparrow$} \\
  \midrule
  W2V2P-LV60 & 15.14 & 51.34 & 79.30 & 62.33 & 6.89 & 67.07 & 71.34 & 69.14 & 7.84 & 67.13 & 77.39 & 71.89 \\
  W2V2P-XLSR53 & 18.97 & 45.64 & 79.94 & 58.10 & 10.40 & 62.01 & 80.57 & 70.08 & 8.71 & 65.64 & 81.53 & 72.73 \\
  MultiIPA & 54.53 & 16.62 & 56.69 & 25.70 & 48.26 & 20.32 & 64.33 & 30.89 & 47.91 & 20.34 & 64.33 & 30.91 \\
  ZIPA-CTC & 11.16 & 57.85 & 78.66 & 66.67 & 6.16 & \textbf{72.32} & 81.53 & \textbf{76.65} & 4.91 & \textbf{77.33} & 84.71 & \textbf{80.85} \\
  ZIPA-CTC-NS & 9.17 & 61.68 & 77.39 & 68.64 & 6.72 & 69.66 & 78.98 & 74.03 & 5.21 & 75.44 & 82.17 & 78.66 \\
  HuPER & 16.00 & 53.64 & \textbf{89.17} & 66.99 & 14.66 & 54.62 & \textbf{86.62} & 67.00 & 15.12 & 55.25 & \textbf{88.85} & 68.13 \\
  PhoneticXeus & 12.20 & 56.98 & 81.85 & 67.19 & 9.45 & 62.87 & 80.89 & 70.75 & 9.23 & 62.84 & 80.25 & 70.49 \\
  BabAR & 74.30 & 9.56 & 39.17 & 15.37 & 71.87 & 10.48 & 42.04 & 16.77 & 43.77 & 21.20 & 59.87 & 31.31 \\
  Azure Speech & \textbf{5.60} & \textbf{70.69} & 78.34 & \textbf{74.32} & 5.41 & 70.66 & 78.98 & 74.59 & 5.41 & 72.08 & 80.57 & 76.09 \\
  Gemini-3.8-Flash & 8.36 & 62.67 & 71.66 & 66.86 & \textbf{3.98} & 53.96 & 23.89 & 33.11 & \textbf{3.02} & 51.58 & 15.61 & 23.96 \\
  GPT-Audio-1.5 & 14.24 & 49.67 & 71.97 & 58.78 & 6.33 & 66.18 & 57.96 & 61.80 & 5.37 & 65.68 & 49.36 & 56.36 \\
  Qwen2.5-Omni-7B & 21.39 & 36.70 & 65.92 & 47.15 & 23.36 & 38.22 & 72.29 & 50.00 & 20.51 & 39.60 & 69.75 & 50.52 \\
  Qwen2.5-Omni-7B$_{\mathrm{TIMIT}}$ & 10.05 & 63.62 & 86.31 & 73.24 & 11.29 & 60.44 & \textbf{86.62} & 71.20 & 11.57 & 61.45 & \textbf{88.85} & 72.66 \\
  Audio Flamingo 3 & 25.24 & 37.39 & 79.30 & 50.82 & 65.30 & 12.93 & 48.73 & 20.44 & 93.79 & 5.22 & 25.80 & 8.68 \\
  Audio Flamingo 3$_{\mathrm{TIMIT}}$ & 13.73 & 56.69 & 86.31 & 68.43 & 13.22 & 55.96 & 83.76 & 67.09 & 18.80 & 46.74 & 82.17 & 59.58 \\
  \bottomrule
  \end{tabular}%
  }
    \caption{Target-word evaluation on the Synthetic Speech Dataset across isolated-word, sentence, and passage contexts. The evaluation is only conducted within the target-word span, including insertions at both word boundaries.}
  \label{tab:synthetic-context-ped-results}
  \end{table*}

\subsection{Prompts}
\label{sec:ALMSpec-prompt}

Qwen2.5-Omni-7B, Audio Flamingo 3, Gemini, and GPT-Audio receive the same system prompt and user prompt, and fine-tuning uses the same prompt as evaluation (see Figure~\ref{fig:ALMSpec-prompt}). When scripted, a short instruction and the intended word and canonical phones follow the user prompt. Qwen2.5-Omni-7B and Audio Flamingo 3 decode greedily with up to 512 new tokens, GPT-Audio uses temperature 0 with up to 512 output tokens, and Gemini uses a low thinking level with up to 4,096 output tokens. All results in the main paper are unscripted unless marked otherwise.

\subsection{Output Parsing}
\label{sec:ALMSpec-parsing}

For Qwen2.5-Omni-7B, Audio Flamingo 3, and Gemini, we strip whitespace and any Markdown code fence around the response and read it as JSON. We take the value under \texttt{transcription}, or the only value if the object has a single key. Extra text after the JSON is ignored, but text before it invalidates the response. A response with no JSON is kept only if it is a plain string of phones, and a response cut off partway through the JSON is recovered when possible. Audio Flamingo 3 sometimes answers in ARPAbet, which we convert to IPA. Gemini and GPT-Audio are called with structured output, which forces a JSON answer. If a response still cannot be parsed, we score it as an empty prediction, so every reference phone counts as deleted. Phones outside the CMU-39 set are not format errors. They go through the projection in Appendix~\ref{sec:cmu39}, and anything that cannot be mapped is dropped (see Table~\ref{tab:ALMSpec-output-validation} for examples).

\section{Additional Experiments}

\subsection{Scripted vs Unscripted Results}
\label{sec:script}

We also conducted experiments to assess whether providing a canonical script improves or degrades speech model performance. Intuitively, the script may constrain predictions and reduce false rejections, which appear to be a common issue for these models. However, it may also bias predictions toward the canonical pronunciation, making models more lenient and less sensitive to phonetic errors. The scripted and unscripted prompts are shown in Figure~\ref{fig:ALMSpec-prompt}.

As shown in Figure~\ref{fig:script}, the results were mixed. On the authentic evaluation data, providing the script reduced Azure's performance (-11.82) but improved performance for Gemini (+9.38), GPT-Audio (+6.19), and Qwen (+5.14). The scripted Qwen model fine-tuned on TIMIT and synthetic combined (SC) data scored 68.61, outperforming scripted Gemini-3.8-Flash (67.64).

On the acted evaluation data, providing the script reduced performance for Azure (-14.37), GPT-Audio-1.5 (-24.06), and our fine-tuned Qwen model (-10.42). These results suggest that consistently integrating information from the canonical script and the audio remains a challenge requiring further investigation. Nevertheless, the unscripted Qwen model fine-tuned on TIMIT and synthetic word (SW) training data scored 90.06, outperforming scripted Gemini-3.8-Flash (82.19).

\subsection{Analysis Focused on Target Words}

As shown in Table~\ref{tab:synthetic-context-ped-results}, we also evaluated PED performance within the target word span, rather than across the entire sentence or passage, to assess whether detection of the embedded error changes across word, sentence, and passage conditions. These metrics include insertions at both boundaries of the target word.

We found that specialized PR models tend to perform better as audio length increases. Specialized PED models, GPT-Audio-1.5, and our Qwen models maintain relatively stable performance across audio lengths. However, both Gemini-3.8-Flash and AF3 show performance degradation with longer context, particularly Gemini-3.8-Flash. These results suggest that differences in model architecture and pretraining data mixtures may influence how well SLMs perform PED on longer audio.

\section{Phone-Level Error Analysis}
\label{sec:appendix-error-analysis}

\begin{figure*}[!t]
    \centering
    \includegraphics[width=0.49\textwidth]{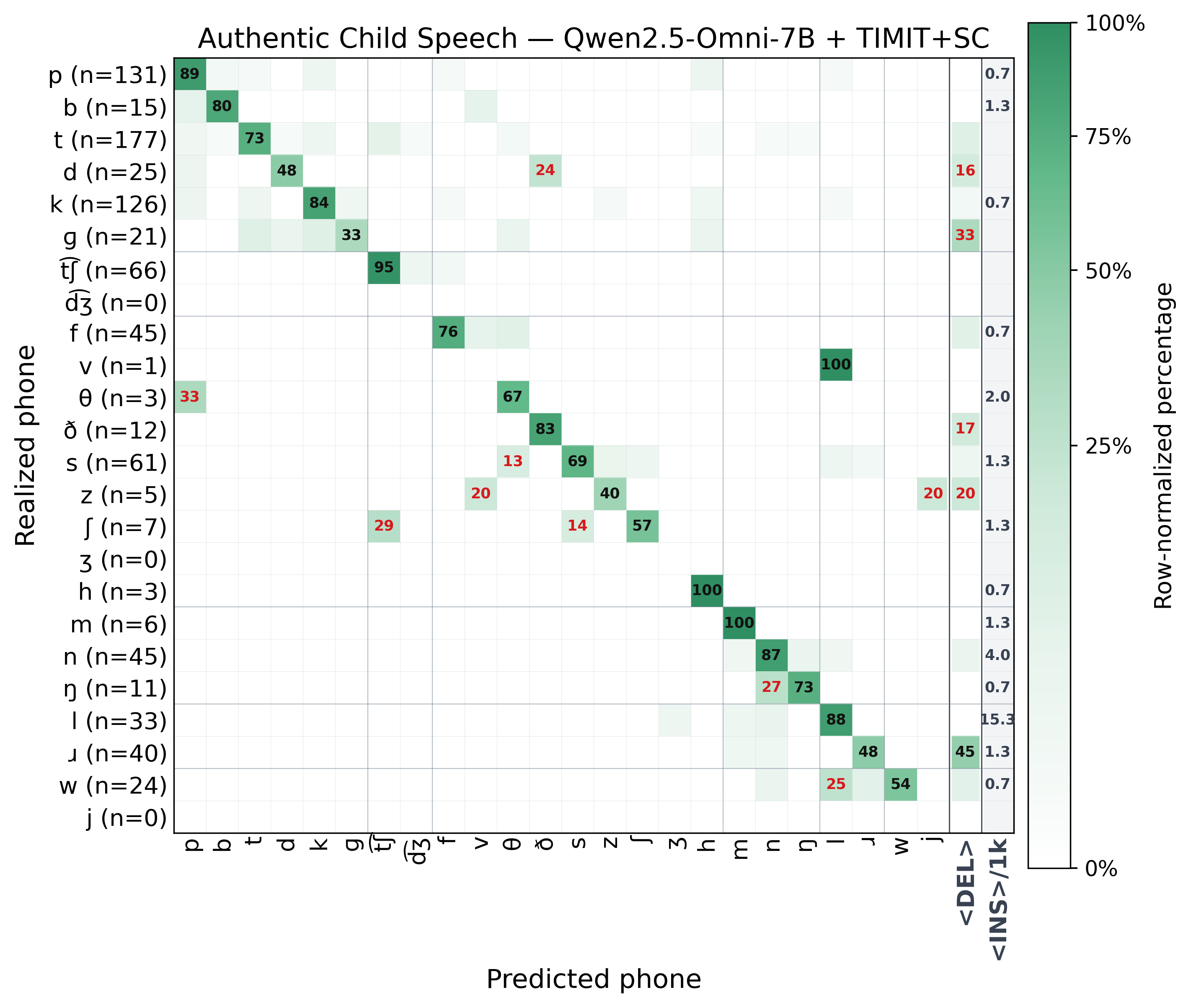}
    \includegraphics[width=0.49\textwidth]{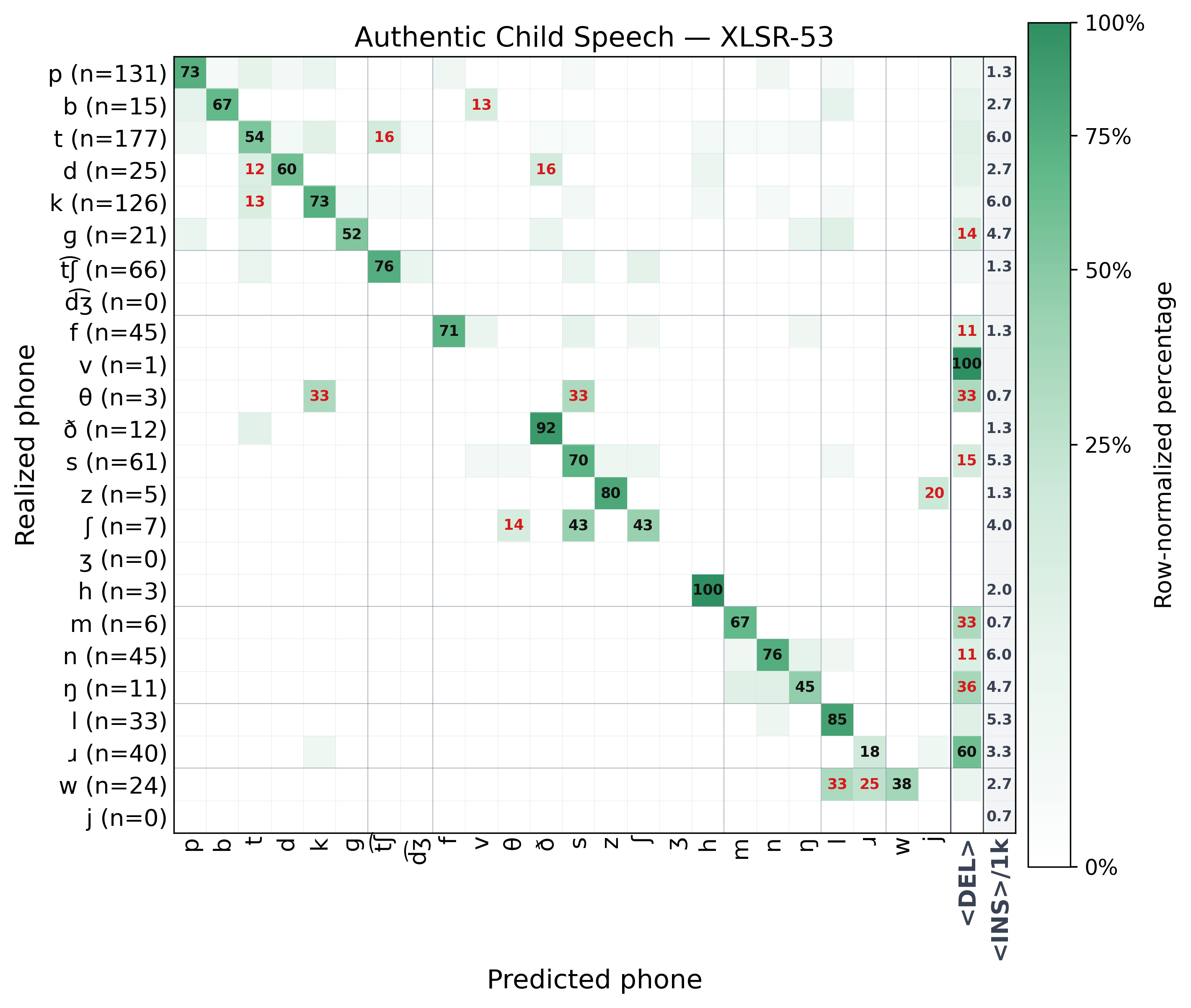}

    \includegraphics[width=0.49\textwidth]{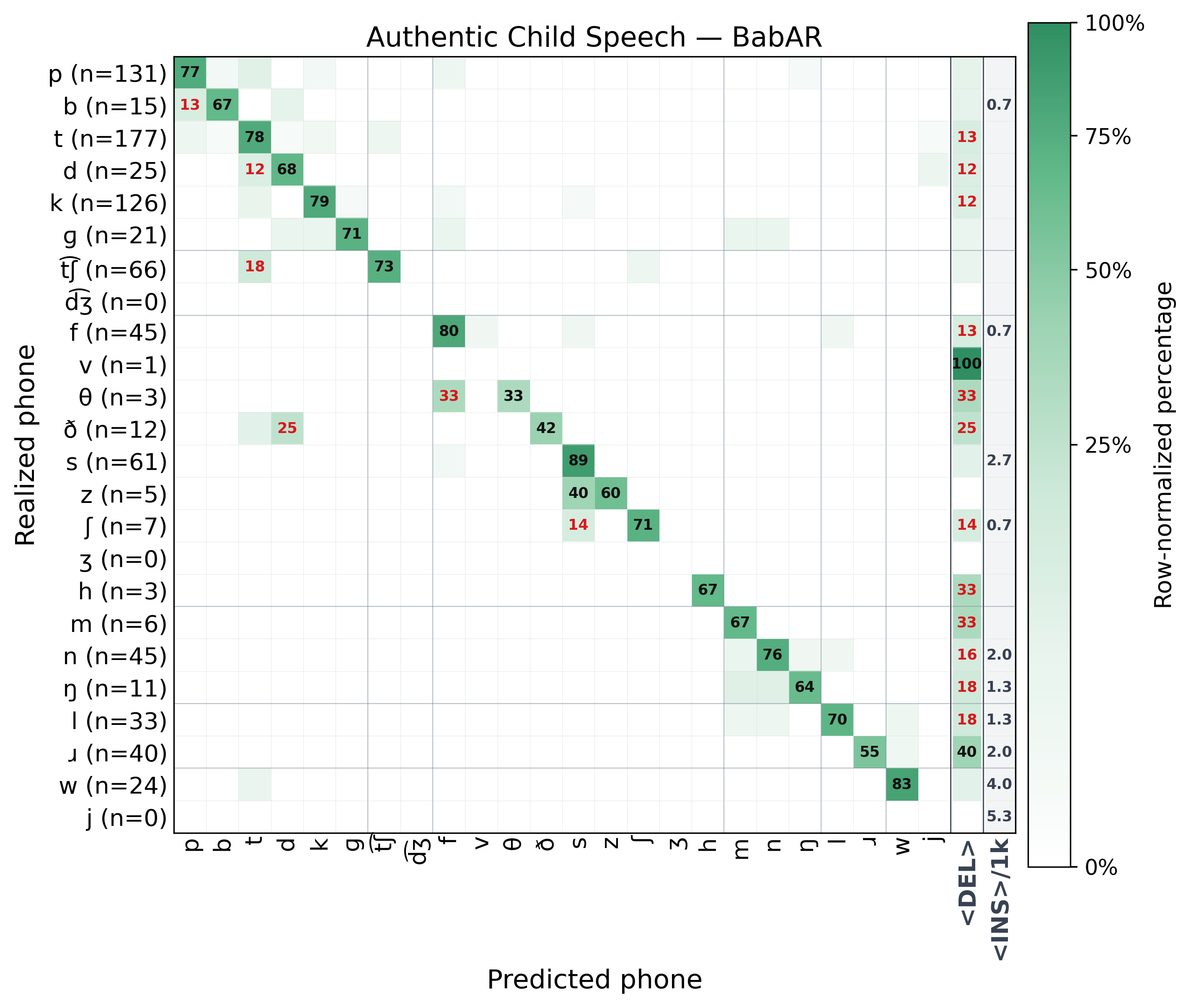}
    \includegraphics[width=0.49\textwidth]{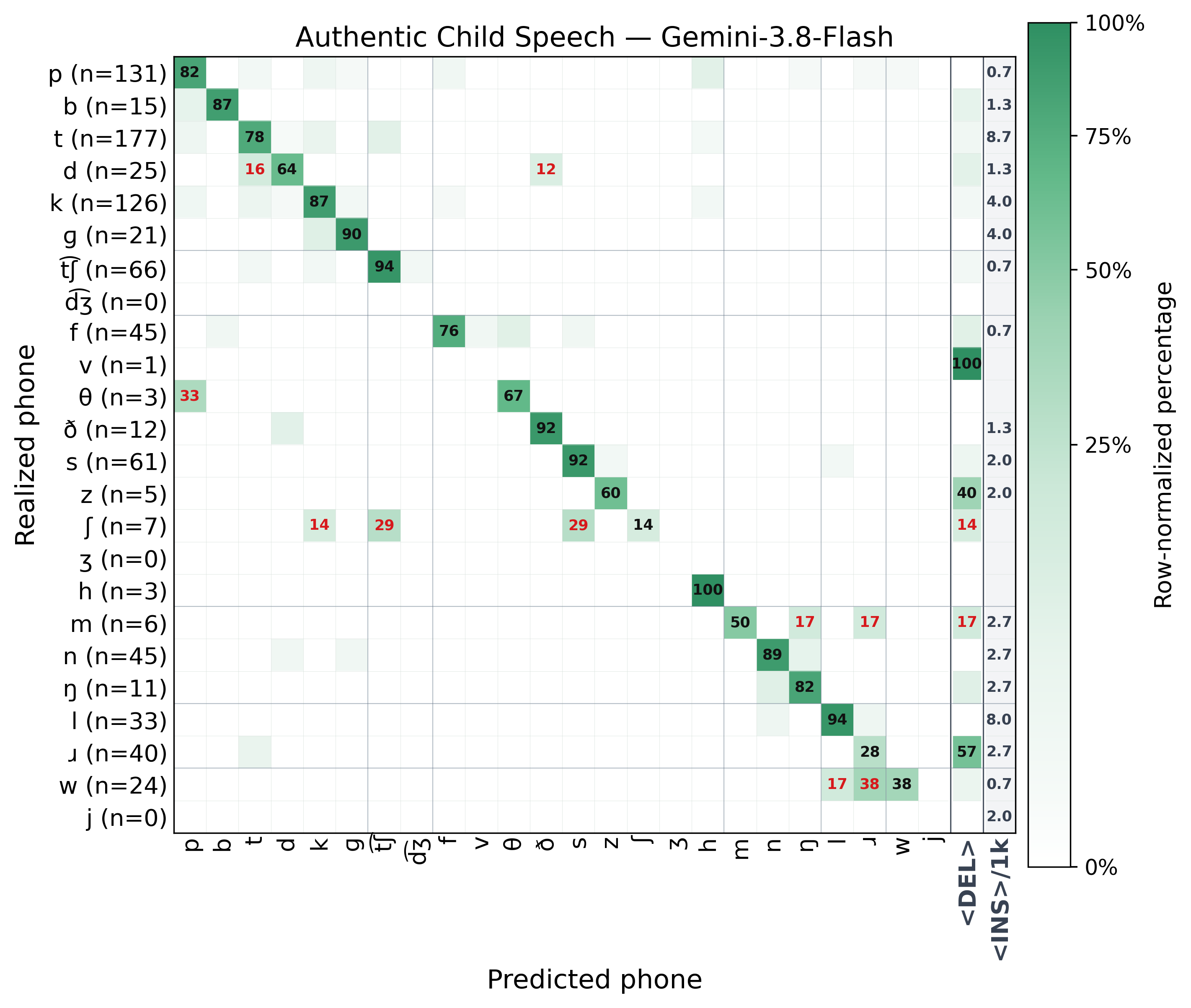}
    \caption{Consonant confusion matrices on Authentic Child Speech for Qwen2.5-Omni-7B + TIMIT+SC (top left), W2V2P-XLSR53 (top right), BabAR (bottom left), and Gemini-3.8-Flash (bottom right).}
    \label{fig:appendix-authentic-consonants}
\end{figure*}

\begin{figure*}[!t]
    \centering
    \includegraphics[width=0.49\textwidth]{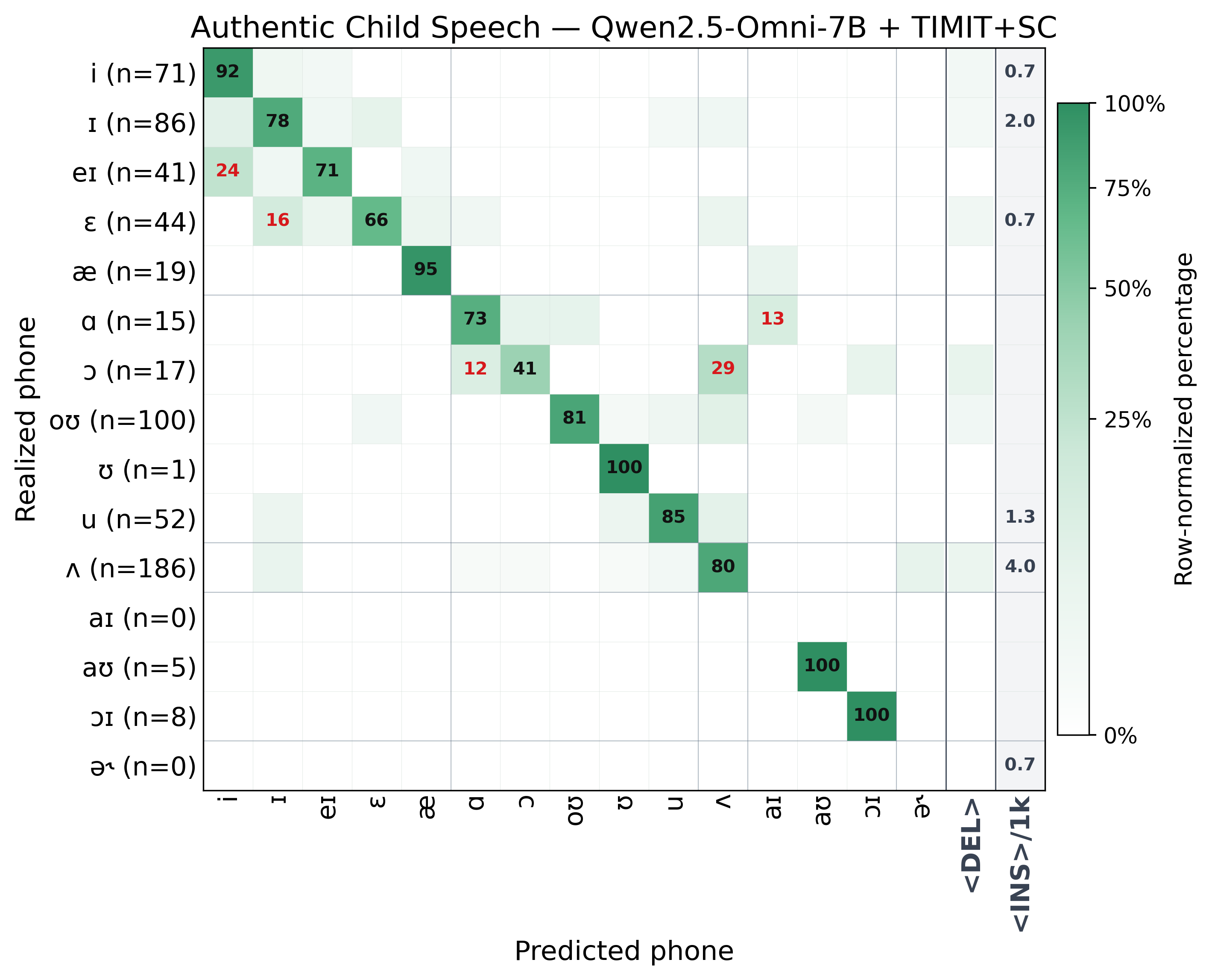}
    \includegraphics[width=0.49\textwidth]{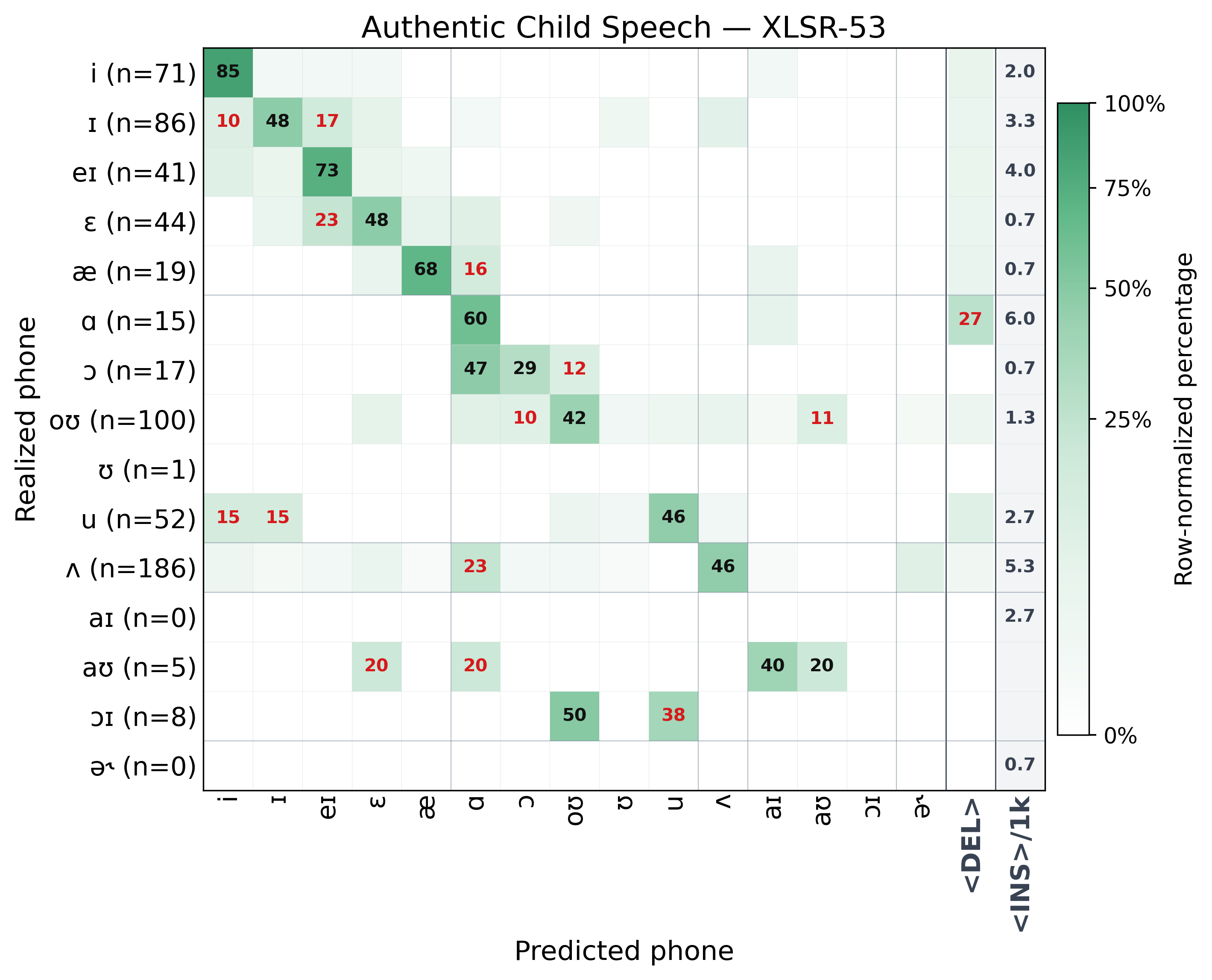}

    \includegraphics[width=0.49\textwidth]{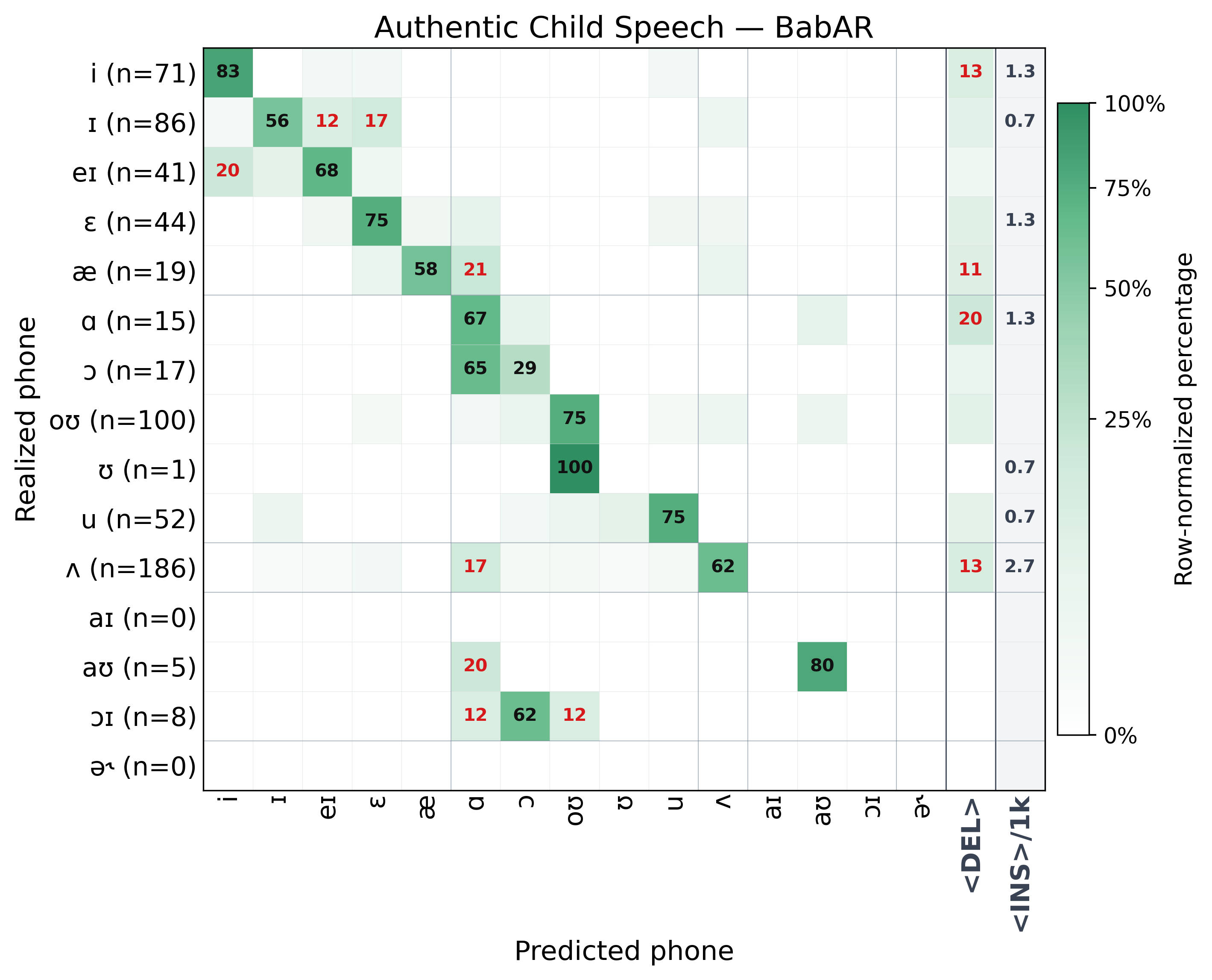}
    \includegraphics[width=0.49\textwidth]{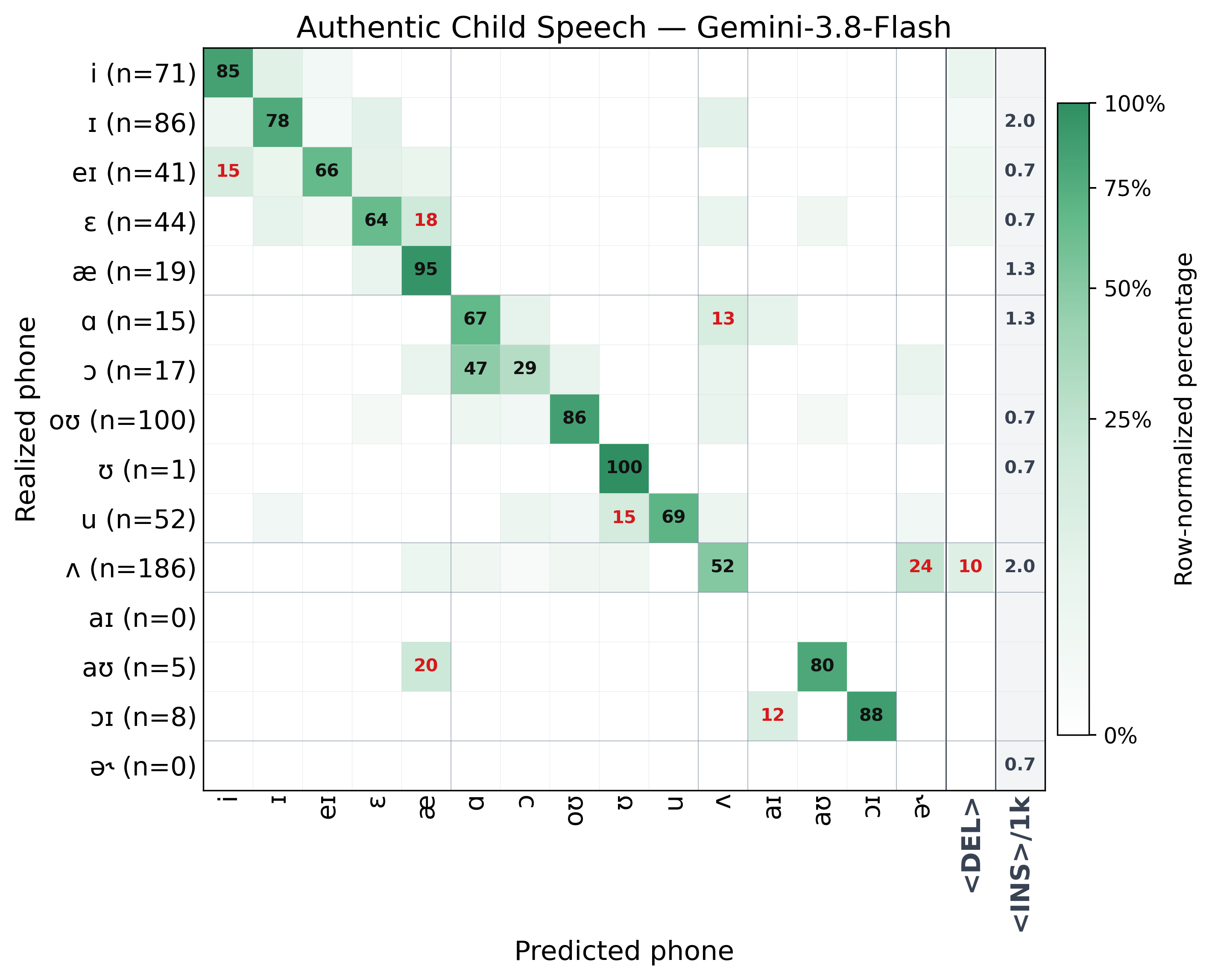}
    \caption{Vowel confusion matrices on Authentic Child Speech for Qwen2.5-Omni-7B + TIMIT+SC (top left), W2V2P-XLSR53 (top right), BabAR (bottom left), and Gemini-3.8-Flash (bottom right).}
    \label{fig:appendix-authentic-vowels}
\end{figure*}

\begin{figure*}[!t]
    \centering
    \includegraphics[width=0.49\textwidth]{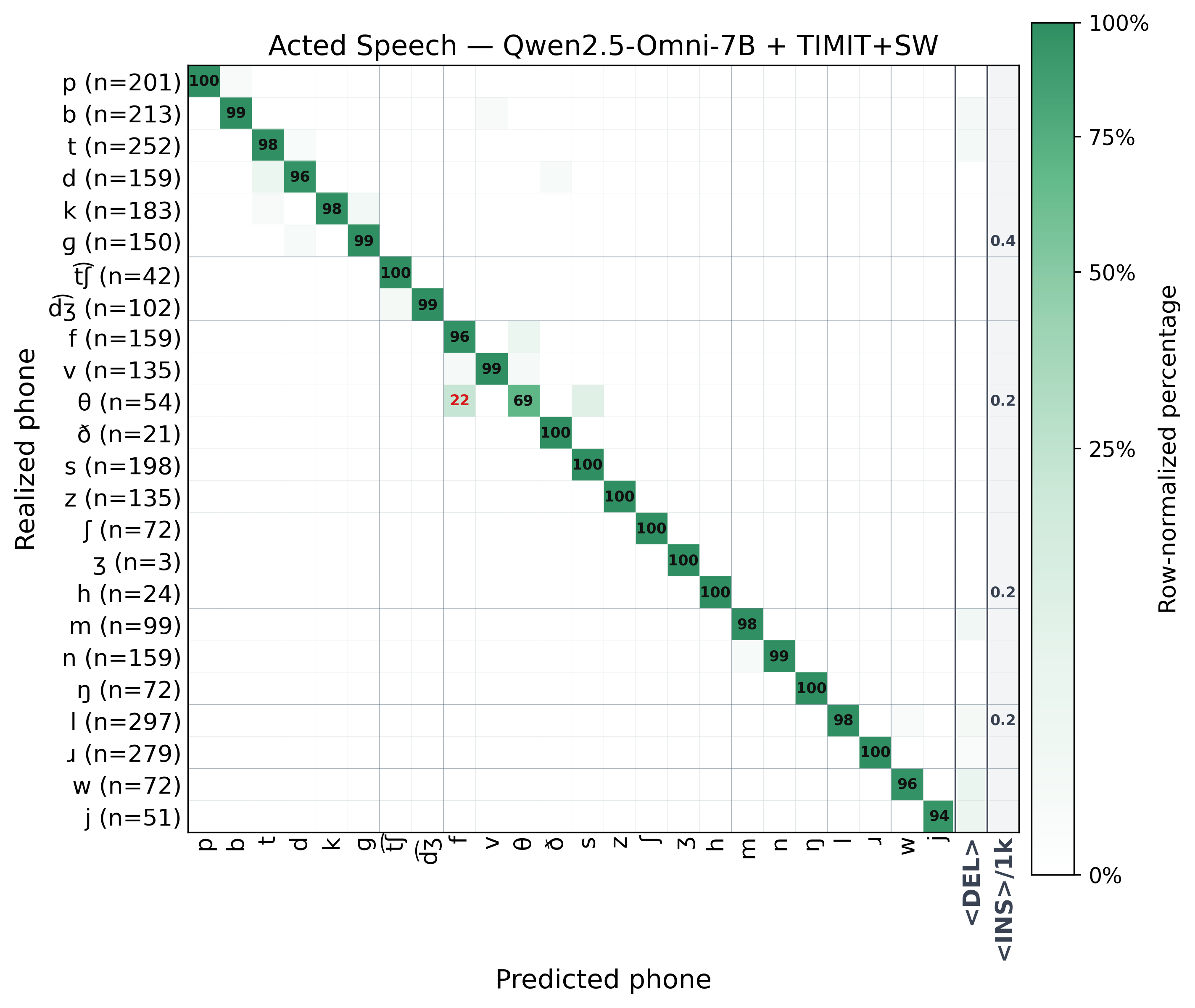}
    \includegraphics[width=0.49\textwidth]{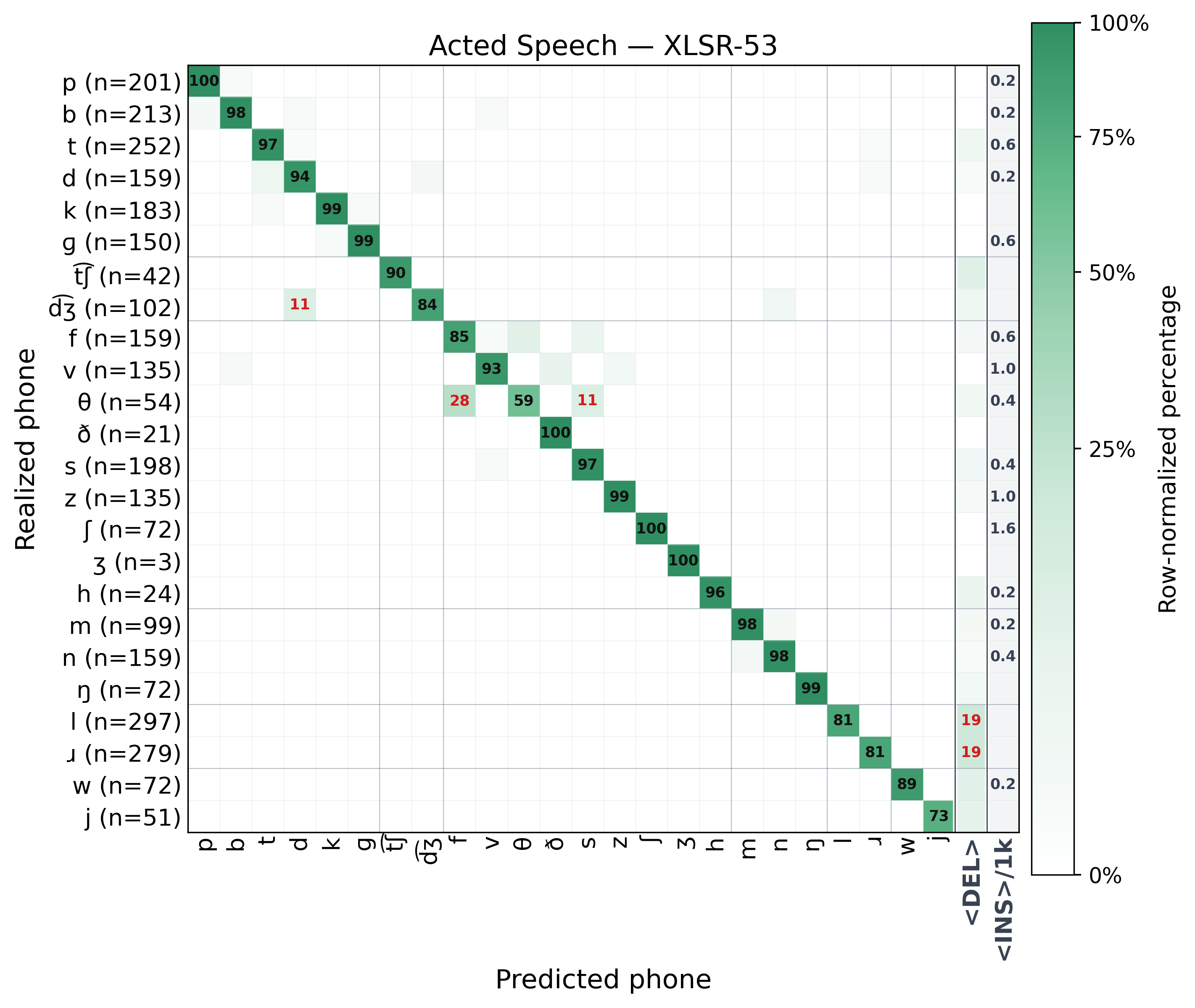}

    \includegraphics[width=0.49\textwidth]{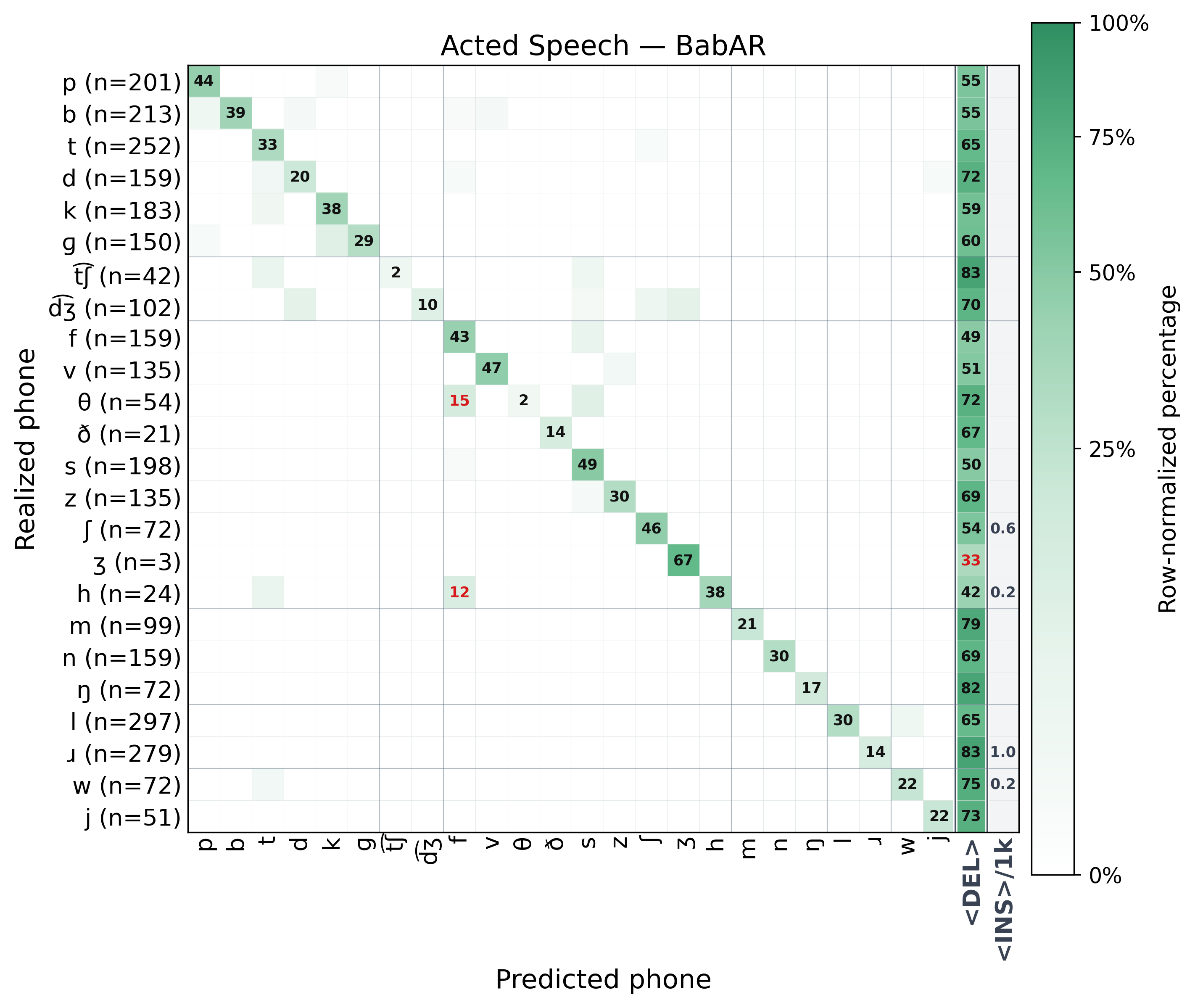}
    \includegraphics[width=0.49\textwidth]{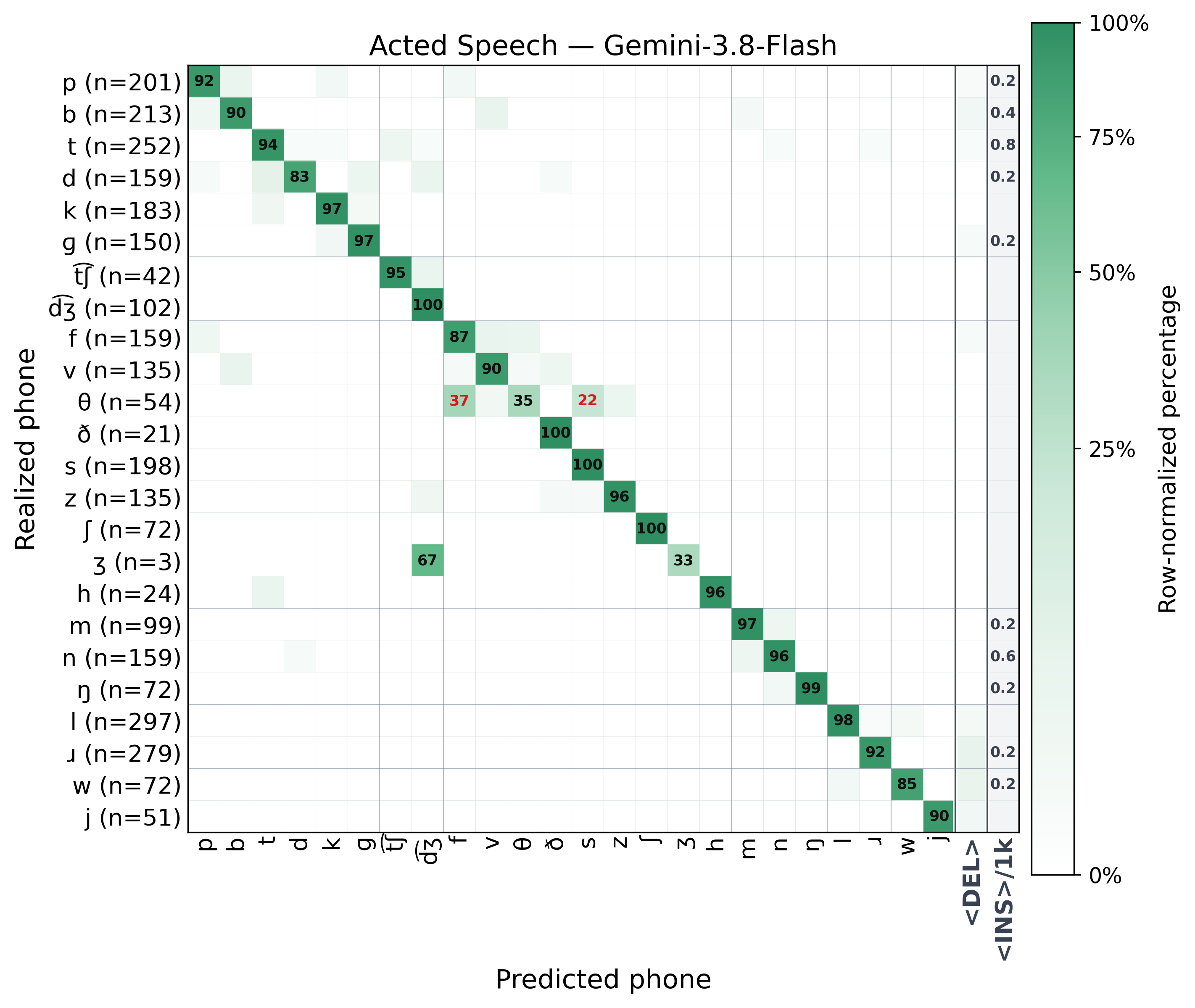}
    \caption{Consonant confusion matrices on Acted Speech for Qwen2.5-Omni-7B + TIMIT+SW (top left), W2V2P-XLSR53 (top right), BabAR (bottom left), and Gemini-3.8-Flash (bottom right).}
    \label{fig:appendix-acted-consonants}
\end{figure*}

\begin{figure*}[!t]
    \centering
    \includegraphics[width=0.49\textwidth]{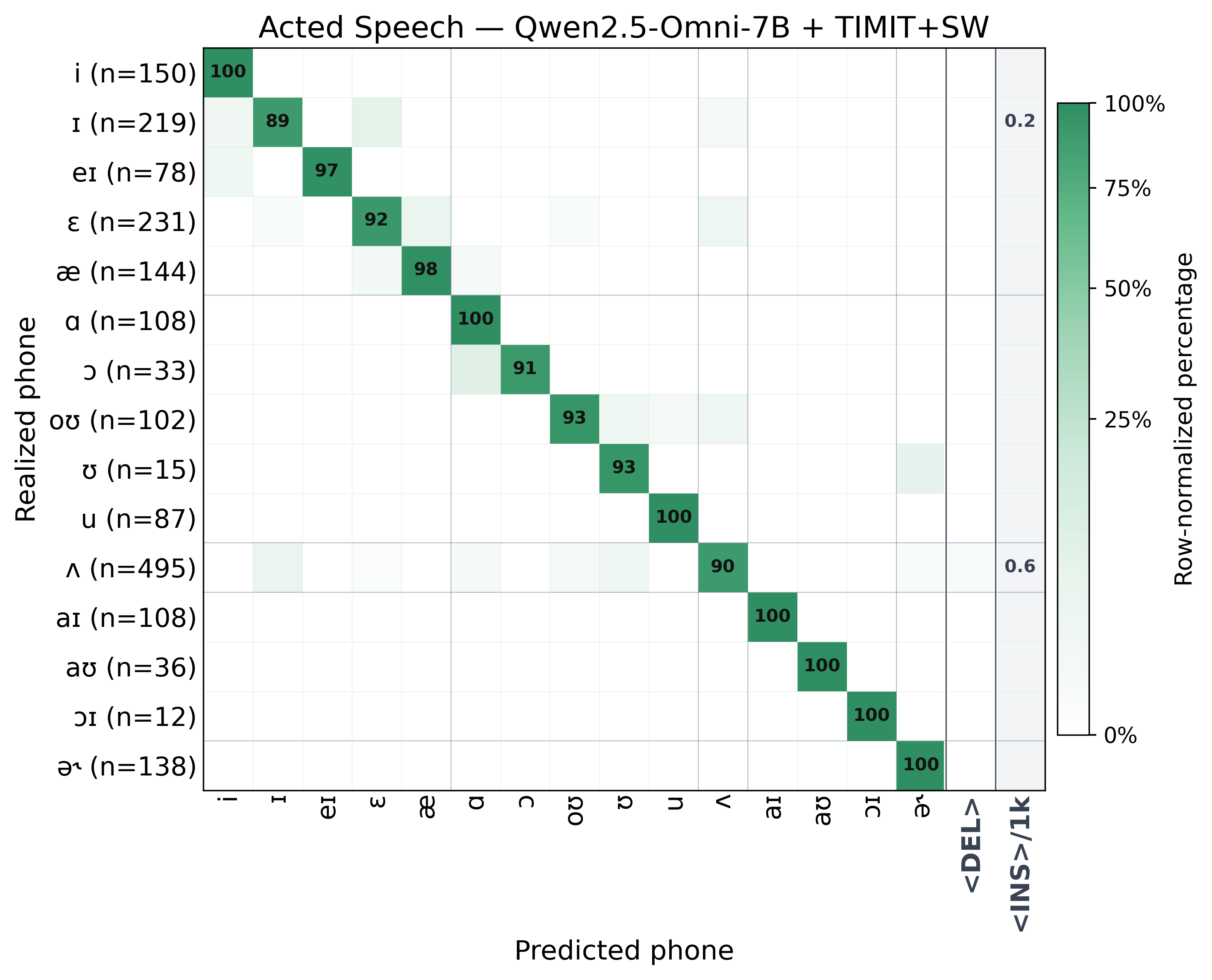}
    \includegraphics[width=0.49\textwidth]{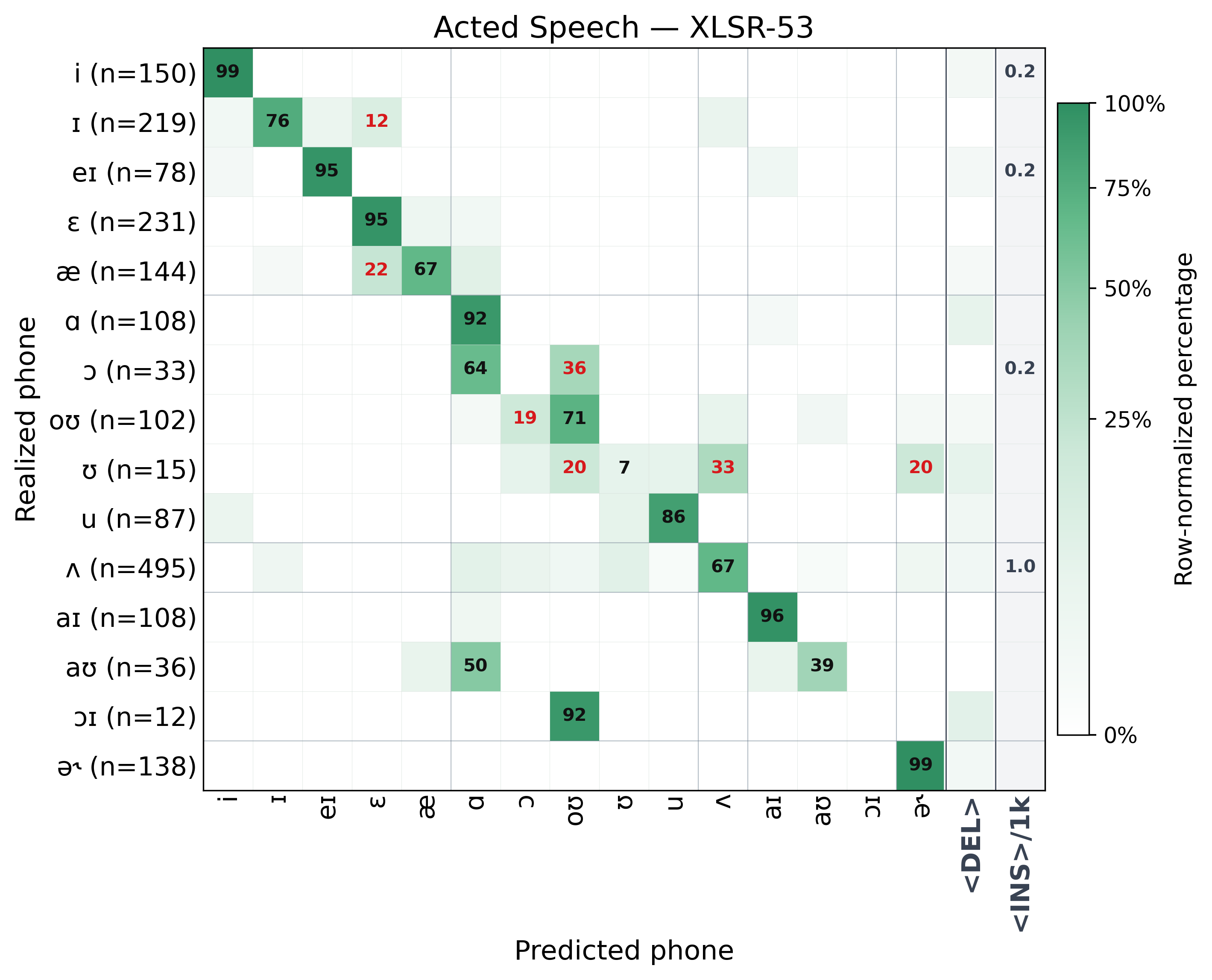}

    \includegraphics[width=0.49\textwidth]{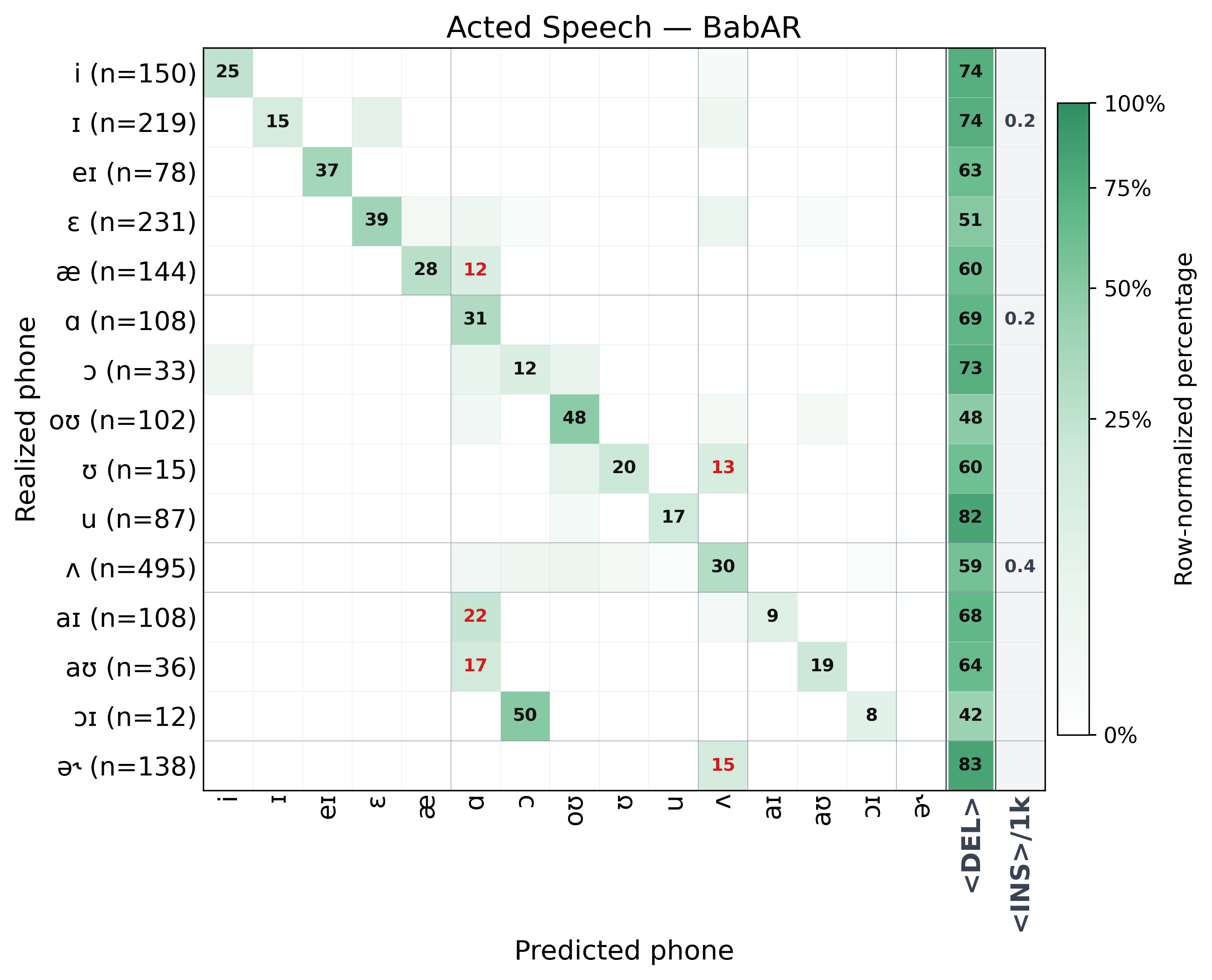}
    \includegraphics[width=0.49\textwidth]{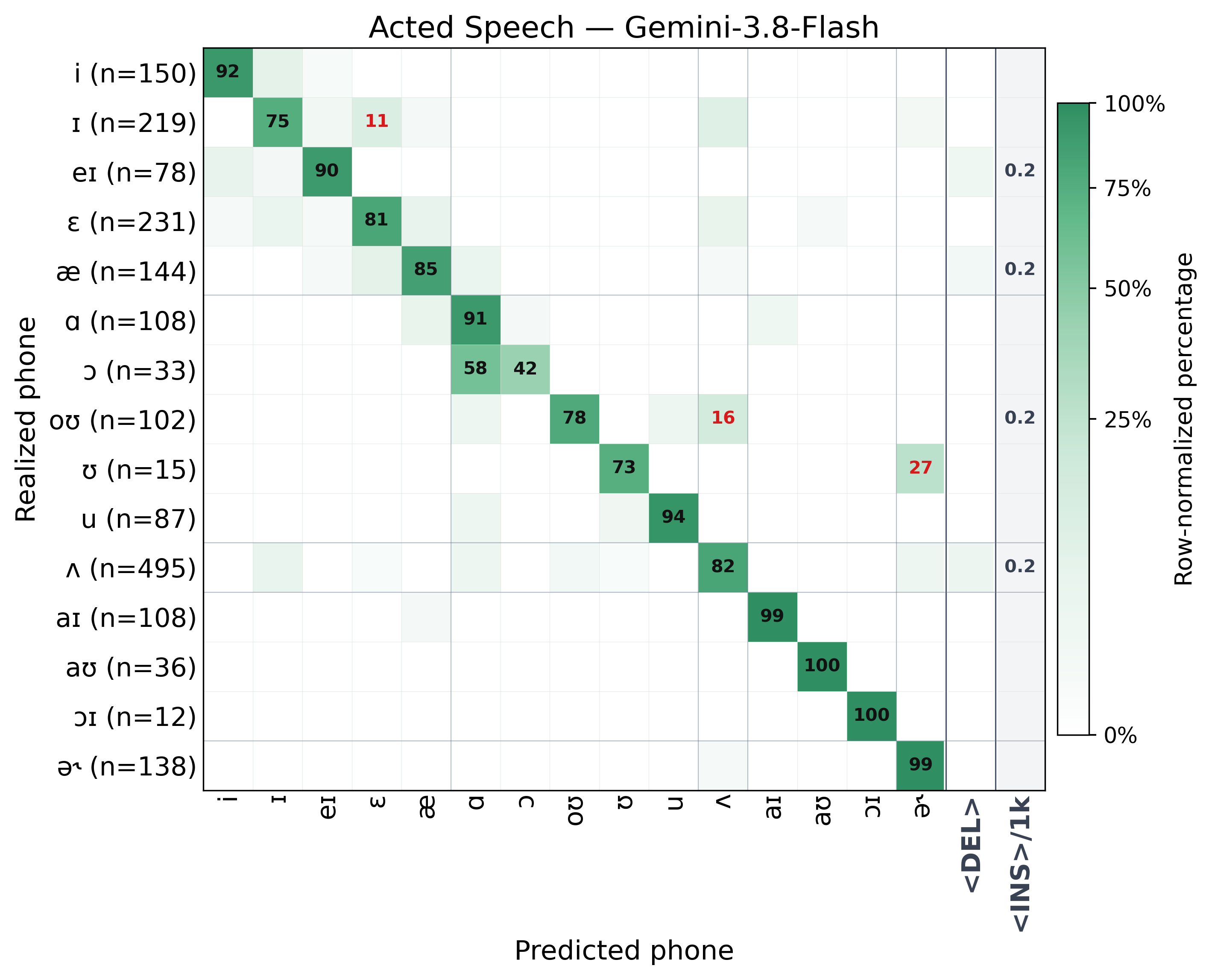}
    \caption{Vowel confusion matrices on Acted Speech for Qwen2.5-Omni-7B + TIMIT+SW (top left), W2V2P-XLSR53 (top right), BabAR (bottom left), and Gemini-3.8-Flash (bottom right).}
    \label{fig:appendix-acted-vowels}
\end{figure*}

For phone-level error analysis, we pick four representative models and analyze their runs on Authentic Child Speech and Acted Speech. We choose our best fine-tuned model (Qwen2.5-Omni-7B + TIMIT+SC on Authentic and + TIMIT+SW on Acted) to represent open-weight speech-language models, W2V2P-XLSR53 to represent open-weight phone recognition models, BabAR to represent child-speech models, and Gemini-3.8-Flash to represent proprietary models. Figures~\ref{fig:appendix-authentic-consonants}--\ref{fig:appendix-acted-vowels} show consonant and vowel confusion matrices. Rows are perceived phones and columns are predicted phones, and each row is normalized by the number of times the phone was perceived ($n$). The \texttt{<DEL>} column shows how often the phone was missed, and the \texttt{<INS>} column shows how often the model inserted that phone where nothing was said, as a share of the same $n$; it is not part of the row's 100\%.

\subsection{Authentic Child Speech}
\label{sec:appendix-error-analysis-authentic}

Qwen+TIMIT+SC gets 77.9\% of the 1,502 perceived phones right (17.1\% substituted, 5.0\% deleted, 41 insertions per 1k phones). Gemini follows with 74.8\% (20.5\% substituted), then BabAR with 71.4\% (11.5\% deleted) and W2V2P-XLSR53 with 58.9\% (33.4\% substituted). All four models often miss \textturnr: 40--60\% of its 40 occurrences are deleted. They also confuse ʌ with nearby vowels. Gemini writes \textipa{@\textrhoticity} for 44 of 186, while W2V2P-XLSR53 and BabAR write ɑ (42 and 32). Three of the four models hear t as \textipa{\t{tS}} (13 to 28 of 177), whereas BabAR does this only 5 times. Gemini and W2V2P-XLSR53 show more insertions (58 and 94 per 1k phones), while Qwen most often inserts l.

\subsection{Acted Speech}
\label{sec:appendix-error-analysis-acted}

Qwen+TIMIT+SW gets 96.7\% of the 5,088 phones right, Gemini 89.9\%, and W2V2P-XLSR53 87.0\%. Most of their remaining errors are vowel substitutions. All three often write ɪ for ʌ (14 to 30 of 495) and ɛ for ɪ (16 to 27 of 219), and all three hear θ as f (12 to 20 of 54). W2V2P-XLSR53 also misses l and \textturnr\ (56 of 297 and 52 of 279). BabAR gets only 28.8\% right because it deletes 64.3\% of phones. BabAR is built to transcribe young children and returns no output for 501 of the 1,188 adult Acted utterances (42\%). On 80 of these recordings, raising the pitch by seven semitones brought its output back for all but 8\%, so we read its Acted results as a domain mismatch rather than poor phone recognition.

\end{document}